\pdfoutput=1

\documentclass[11pt]{article}

\usepackage[]{acl}

\usepackage{times}
\usepackage{latexsym}
\usepackage{colortbl}
\usepackage[T1]{fontenc}

\usepackage[utf8]{inputenc}

\usepackage{microtype}

\usepackage{inconsolata}

\usepackage{subfigure}
\usepackage{amsmath}
\usepackage{amssymb}
\usepackage{mathtools}
\usepackage{amsthm}
\usepackage{booktabs} 
\usepackage{multirow}
\usepackage{graphicx}
\usepackage{hyperref}
\usepackage{url}
\usepackage{pgfplots}
\usetikzlibrary{pgfplots.groupplots}
\usepackage{wrapfig, lipsum, booktabs}
\pgfplotsset{compat=1.3}
\usepackage{tikz}
\usepackage{listings}
\lstdefinestyle{prompt}{
  basicstyle=\ttfamily\scriptsize,
  breaklines=false,
  frame=single,
  columns=fullflexible,
  escapeinside={(*}{*)},
  backgroundcolor=\color{gray!5!white},
}

\usetikzlibrary{patterns}
\usepackage{pifont}
\usepackage[T1]{fontenc}
\usepackage{arydshln}
\usepackage{soul}
\usepackage{hhline}

\usepackage{cleveref}
\crefname{section}{§}{§§}
\Crefname{section}{§}{§§}

\usepackage[ruled]{algorithm2e}
\usepackage{algorithmic} 

\newcommand{\okmark}{{\textbf{\textcolor[rgb]{0.1, 0.5, 0.1}{$\checkmark$}}}}
\newcommand{\ngmark}{{\textbf{\color{red}{\ding{55}}}}}
\definecolor{myblue}{RGB}{215,238,247}
\definecolor{mypink}{RGB}{236,168,169}
\definecolor{arrowblue}{RGB}{0,113,188}
\definecolor{greymodule}{RGB}{242,242,242}

\definecolor{battleshipgrey}{rgb}{0.3, 0.3, 0.3}
\definecolor{brilliantrose}{rgb}{1.0, 0.33, 0.64}
\definecolor{americanrose}{rgb}{1.0, 0.01, 0.24}
\definecolor{jweigreen}{rgb}{0,0.45,0.24}
\definecolor{bluegray}{rgb}{0.1, 0.1, 0.4}
\definecolor{ao(english)}{rgb}{0.0, 0.5, 0.0}
\definecolor{blanchedalmond}{rgb}{1.0, 0.92, 0.8}
\definecolor{atomictangerine}{rgb}{1.0, 0.6, 0.4}
\definecolor{chocolate(web)}{rgb}{0.82, 0.41, 0.12}
\definecolor{bananayellow}{rgb}{1.0, 0.88, 0.21}
\definecolor{goldenbrown}{rgb}{0.6, 0.4, 0.08}
\definecolor{aliceblue}{rgb}{0.94, 0.97, 1.0}
\definecolor{beige}{rgb}{0.96, 0.96, 0.86}
\definecolor{babyblue}{rgb}{0.54, 0.81, 0.94}
\definecolor{camel}{rgb}{0.76, 0.6, 0.42}
\definecolor{cinnamon}{rgb}{0.82, 0.41, 0.12}

\definecolor{mygrey}{RGB}{202,202,202}

\definecolor{myblue}{RGB}{145,179,206}

\definecolor{myorange}{RGB}{247,164,68}

\definecolor{mypink}{RGB}{241,202,196}

\definecolor{mygreen}{RGB}{143,185,67}

\usepackage{pgfplots}
\usetikzlibrary{pgfplots.groupplots}
\pgfplotsset{compat=1.3}
\usepackage{pifont}
\pgfplotsset{compat=1.11,
    /pgfplots/ybar legend/.style={
    /pgfplots/legend image code/.code={%
       \draw[##1,/tikz/.cd,yshift=-0.25em]
        (0cm,0cm) rectangle (7pt,0.8em);},
   },
}

\title{Generalizable Chain-of-Thought Prompting in Mixed-task Scenarios with Large Language Models}
\author{Anni Zou$^\dagger$, Zhuosheng Zhang$^\dagger$, Hai Zhao$^\dagger$, Xiangru Tang$^\ddagger$\\
$^\dagger$Shanghai Jiao Tong University, $^\ddagger$Yale University\\
\texttt{\{annie0103,zhangzs\}@sjtu.edu.cn,zhaohai@cs.sjtu.edu.cn,}\\
\texttt{xiangru.tang@yale.edu}\\
}

\begin{document}
\maketitle
\begin{abstract}
Large language models (LLMs) have unveiled remarkable reasoning capabilities by exploiting chain-of-thought (CoT) prompting, which generates intermediate reasoning chains to serve as the rationale for deriving the answer. 
However, current CoT methods either simply employ general prompts such as \emph{Let's think step by step}, or heavily rely on pre-defined task-specific demonstrations to attain preferable performances, thereby engendering an inescapable gap between performance and generalization. 
To bridge this gap, we propose \textbf{GeM-CoT}, a \underline{\textbf{Ge}}neralizable CoT prompting mechanism in \underline{\textbf{M}}ixed-task scenarios where the type of input questions is unknown. 
GeM-CoT first categorizes the question type and subsequently samples or constructs demonstrations from the corresponding data pool in an automatic pattern. 
With this technical design, GeM-CoT simultaneously enjoys superior generalization capabilities and remarkable performances on 10 public reasoning tasks and 23 BBH tasks. 
\end{abstract}

\section{Introduction}
Large language models (LLMs) \citep{few-shot, scao2022bloom, thoppilan2022lamda, chowdhery2022palm, touvron2023llama, openai2023gpt4} have exhibited commendable capabilities on complex reasoning by virtue of chain-of-thought (CoT) prompting \citep{wei2023chainofthought}. CoT prompting entails the generation of intermediate reasoning chains that serve as the rationale before deriving the answer.

Current CoT prompting methods predominantly fall into two categories, which we dub as \emph{General Zero-Shot-CoT} and \emph{Specific Few-Shot-CoT}, respectively. The former leverages general trigger prompts such as \emph{Let's think step by step} and appends them directly to the input question, aiming to summon up the step-by-step reasoning potential from LLMs \citep{zero-shot, yang2023large}. The latter provides task-specific input-output pairs as in-context demonstrations and puts them before the input question, for the purpose of instructing LLMs to carry out multi-step reasoning with elaborately selected demonstrations \citep{liu-etal-2022-makes,wei2023chainofthought, zhang2023automatic}.

\begin{figure}
    \centering
    \includegraphics[width=0.5\textwidth]{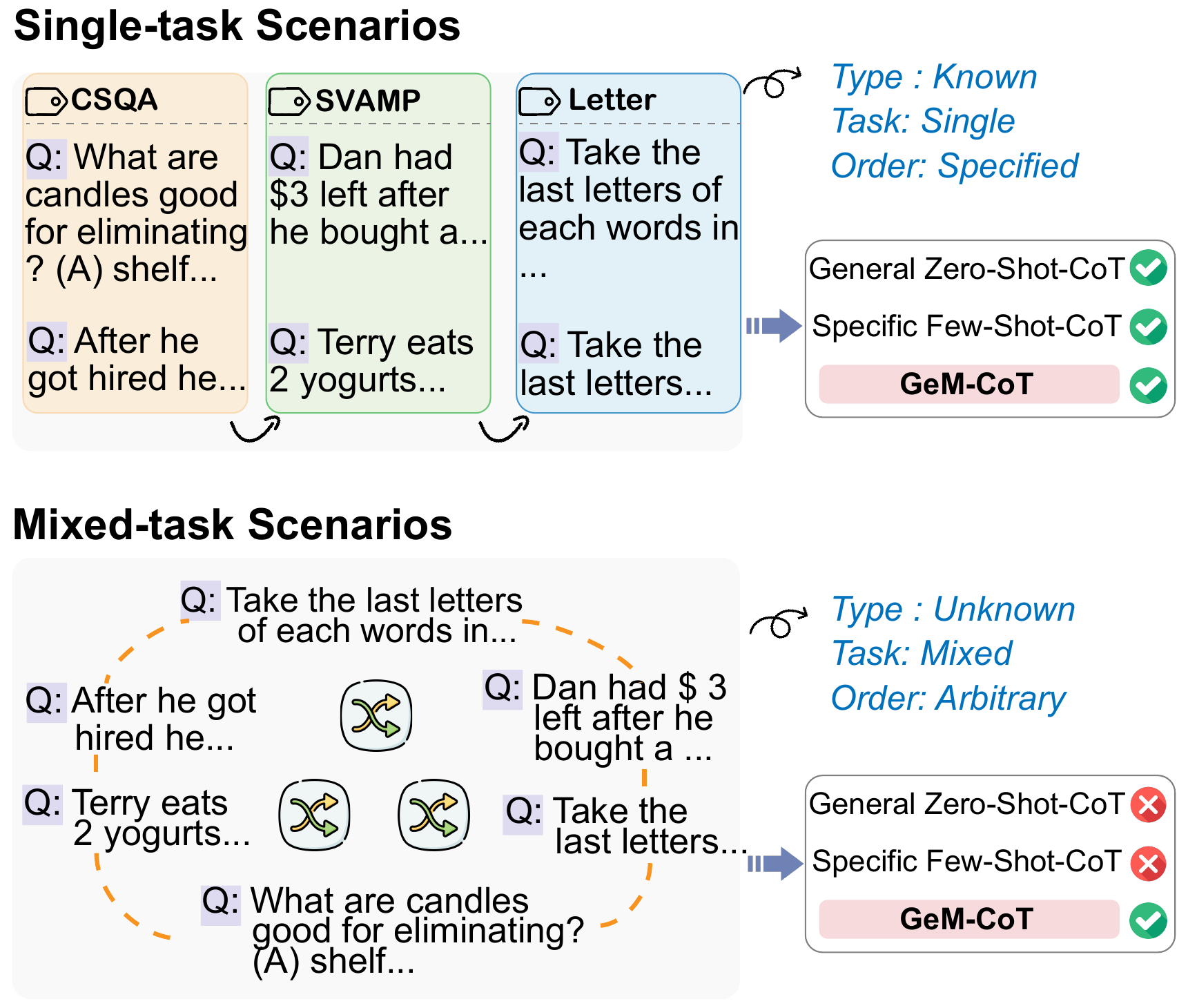}
    \caption{Comparison of conventional \emph{single-task scenarios} and our concerned setting: {\textbf{mixed-task scenarios}}. There are three major characteristics of mixed-task scenarios: (i) the type of any incoming question is unknown; (ii) the input data comes from a set of mixed tasks; (iii) the questions come in an arbitrary order.}
    \label{fig:intro_scenario}
    \vspace{-.4cm}
\end{figure}

Briefly, there are two major limitations in previous studies. On one hand, the \emph{General Zero-Shot-CoT} pattern is endowed with favorable generalization ability as it does not need any task-related demonstrations, but it often pales in terms of performance when compared with the few-shot pattern. On the other hand, the \emph{Specific Few-Shot-CoT} pattern heavily leans on task-specific demonstrations to attain superior performances, yet fails to bear on decent generalization ability. Although recent works have made progress by either mitigating manual labor \citep{zhang2023automatic} or promoting the quality of demonstrations \citep{arora2022ask, diao2023active}, all of them rest on the task-associated perspective thus far.

Nevertheless, in practical applications, LLMs tend to confront situations of mixed types of questions (Figure \ref{fig:intro_scenario}), where each question is not clearly pre-identified which task it belongs to. Under these circumstances, it is neither reasonable to improvise several task-related examples by hand nor possible to manually search for which task it refers to, not to mention that the question encountered in actual cases is not even from a pre-defined set of tasks. Besides, naive use of general trigger prompts may result in performance degradation as the lack of templated rationales often leads to spurious reasoning steps \citep{wan-etal-2023-better}. As a result, there exists an inescapable gap between performance and generalization in our concerned realistic mixed-task scenarios.\footnote{Detailed exploration will be provided in Section \ref{sec:challenge}.} To alleviate this gap, a potential strategy is to explore the trade-off area between generality and performance while ensuring certain practicality.

This work presents \textbf{GeM-CoT}: a \underline{\textbf{Ge}}neralizable CoT prompting mechanism in \underline{\textbf{M}}ixed-task scenarios where the type of input questions is unknown. 
GeM-CoT first routes the input question to different paths based on whether it can successfully match to a demo pool that is pre-constructed and continuously updated. On one hand, for a successful match, it fetches demonstrations of the matched type from the demo pool and performs a final inference to acquire the answer. On the other hand, when a match fails, it derives the answer through zero-shot reasoning and then stores in the data cache. Afterward, it updates the cache by conducting density-based clustering on the questions within and automatically constructing diverse demonstrations for data in a certain cluster that meets the requirements. The corresponding generated demonstrations are returned to the demo pool for subsequent inference.

We conduct experiments on 10 reasoning tasks covering arithmetic reasoning, commonsense reasoning, and symbolic reasoning. Besides, we further validate the stability and generalization of GeM-CoT on 23 BBH datasets. Experimental results show that GeM-CoT simultaneously enjoys superior generality and remarkable performances. 

Our contributions are summarized as follows:

(i) To the best of our knowledge, our work pioneers a novel setting of mixed-task scenarios, which has significant practical application values.

(ii) We propose a generalizable CoT prompting mechanism in mixed-task scenarios, which not only bridges the gap between performance and generalization but also unearths their in-between mutual synergy by gaining performance improvements in sync with achieving generality.

 (iii) Experimental results on a total of 33 datasets demonstrate the impressive performance and superior generality of our approach.

\section{Related Work}
In this section, we discuss two lines of research which are key to our work: CoT prompting and cross-task generalization. 

\subsection{Chain-of-thought Prompting}
Recently, CoT prompting methods have pushed the multi-step reasoning abilities of LLMs to a remarkable aptitude by eliciting them to generate intermediate reasoning chains before deriving the final answer \citep{wei2023chainofthought}.

Currently, there are two flavors of research in CoT prompting: \emph{General Zero-Shot-CoT} \citep{zero-shot} and \emph{Specific Few-Shot-CoT} \citep{wei2023chainofthought}. The former merely appends a \underline{\emph{general}} prompt to the input question, wheras the latter leverages several task-\underline{\emph{specific}} input-output pairs as reasoning demonstrations and inserts them before the test question.

\paragraph{General Zero-Shot-CoT.}
LLMs have proven to be competent zero-shot reasoners by \citet{zero-shot}, which has greatly broadened the generalizability of CoT techniques and liberated the need to prepare task-specific examples in advance. While benefiting from its task-agnostic property, it often fails to excel at performance in comparison with its few-shot rivals \citep{wei2023chainofthought, zhang2023automatic}. In order to further boost the performance, recent works have laid emphasis on the optimization of triggering prompts \citep{zhou2022large,yang2023large}. In their work, LLMs are employed as optimizers, and new prompts are progressively generated based on the past optimization history. Despite the augmented performance, the optimization process for prompts reverts to a task-specific problem, and for unseen test questions in real-world circumstances, it may not be advisable to optimize prompts on the fly.

\paragraph{Specific Few-Shot-CoT.}
Owing to the well-crafted in-context demonstrations, Few-Shot-CoT achieves preferable performance, which consequently extends to a plethora of studies focusing on improvements upon it. According to the period of improvement, these studies are grouped into three categories: (i) pre-reasoning pattern; (ii) peri-reasoning pattern; and (iii) post-reasoning pattern. 

For the pre-reasoning pattern, current research attends to either alleviating manual labor when selecting demonstrations \citep{zhang2023automatic,wan-etal-2023-better}, or promoting demonstration quality \citep{creswell2022selectioninference, madaan2022text,arora2022ask,diao2023active,wang2023meta}. 
For the post-reasoning pattern, recent studies concentrate on fine-grained reasoning processes such as problem decomposition \citep{zhou2023leasttomost,press2022measuring}. 
For the post-reasoning pattern, related works principally enhanced the performance by verification \citep{weng2023large, lyu2023faithful} or ensemble-like methods \citep{wang2023selfconsistency, li2023making, wang2022rationaleaugmented, yoran2023answering}. 

However, the aforementioned works, which mainly hinge on task-associated demonstrations, fail to step outside the task-specific framework to pursue generalizability. In turn, there is an upper bound to the performance that a general Zero-Shot-CoT method can achieve, thus leading the current CoT prompting to a dilemma. Our work, in contrast, manages to find a way out of this dilemma by intuitively carrying out a routing mechanism, making our proposed GeM-CoT applicable in realistic mixed-task scenarios.

\subsection{Cross-task Generalization}
Cross-task generalization has been a long-standing research goal in natural language processing (NLP). The conventional pre-training and fine-tuning paradigm gains a foothold by pre-training on a large corpus of text to capture general knowledge and fine-tuning on specific tasks to acquire specific knowledge. Beyond this primitive paradigm, post pre-training and multi-task learning \citep{yu2022dictbert, zhang-zhao-2021-structural, liu2019multitask, zhang-etal-2022-task} encourage further advancements in this research area. More recent works such as ExT5 \citep{aribandi2022ext5}, T0 \citep{sanh2022multitask}, and FLAN \citep{wei2022finetuned} strived to convert a variety of tasks into an identical text-to-text format, so that models can be trained on those tasks jointly. LoraHub \citep{huang2023lorahub} leveraged the composability of LoRA (Low-Rank Adaption of LLMs) modules to promote the task generalization ability of LLMs. Our work, however, manages to effectuate task generalization through timely and user-friendly ICL without any training.

\section{Towards Generalizable CoT in Mixed-task Scenarios}
In this section, we first define the concept of mixed-task scenarios and then present preliminary experiments to understand the challenge.

\subsection{Concept of Mixed-task Scenarios}\label{sec:bg}
Existing studies \citep{wei2023chainofthought} commonly assume that the type of questions fed to the model is known and conduct each set of evaluations on the questions from the same dataset, which is regarded as the single-task scenarios. However, a more realistic setting lies in \textbf{mixed-task scenarios} where the type of input questions is unknown and they come in an arbitrary manner. A comparison with the single-task scenarios is presented in Table \ref{tab:concept}.

\begin{table}[htb]
 \centering
 \small
 
 \setlength{\tabcolsep}{4.5pt}
 {\begin{tabular}{lccc}
  \toprule
            \multirow{2}{*}{Setting} & Unknown & Mixed & Arbitrary \\ 
            & Type & Source & Order \\ 
            \midrule
            Single-task Scenarios & \ngmark & \ngmark & \ngmark\\
            \textbf{Mixed-task Scenarios} & \okmark & \okmark & \okmark\\
  \bottomrule
 \end{tabular}
 }
  \caption{Concept of \textbf{mixed-task scenarios}, which is more common in real-world situations.}\label{tab:concept}    
\end{table}

Mixed-task scenarios have three main characteristics: (i) the type of any incoming question is unknown; (ii) the input data comes from a set of mixed tasks; (iii) the questions come in an arbitrary order. Such a setting is of pivotal importance because the specific task source of an incoming question is usually unavailable in many real-world applications.

\subsection{Challenge of Mixed-task Scenarios}\label{sec:challenge}
In the first place, we set up the mixed-task scenarios by adopting questions from ten reasoning tasks following \citet{zero-shot} and \citet{zhang2023automatic}. We shuffle all the questions and sample 100 examples to mimic their mixed and arbitrary pattern. We initially adopt two vanilla methods: Zero-Shot-CoT and Few-Shot-CoT,\footnote{We leverage ICL demonstrations from \citet{wei2023chainofthought} and refer them as \emph{gold demos}.} the latter assuming a known dataset source for the input question, which cannot be applied to the mixed-task scenarios, but only serves a hypothetical upper bound for reference.

\begin{table}[htb]
 \centering\small
 
 \setlength{\tabcolsep}{6pt}
 {\begin{tabular}{lcl}
  \toprule
            \multirow{2}{*}{Method} & Mixed-task & \multirow{2}{*}{Accuracy} \\ 
            & Scenarios &  \\ 
            \midrule
            Few-Shot-CoT (w/ gold) & \ngmark & 78.0\\
            \midrule
            \textbf{\emph{zero-shot setting}} & &\\
            \quad Zero-Shot-CoT & \okmark & 66.0 ($\downarrow$ 12.0)\\
            \hdashline
            \textbf{\emph{few-shot setting}} & &\\
            \quad w/ varied \& single & \okmark & 26.0 ($\downarrow$ 52.0)\\
            \quad w/ varied \& mixed & \okmark & 20.0 ($\downarrow$ 58.0)\\
            \quad w/ fixed \& single & \okmark & 27.0 ($\downarrow$ 51.0)\\
            \quad w/ fixed \& mixed & \okmark & 19.0 ($\downarrow$ 59.0)\\
  \bottomrule
 \end{tabular}
 }
  \caption{Results with initial attempts showing the challenge of mixed-task scenarios.}\label{tab:challenge}
\end{table}

As seen in Table \ref{tab:challenge}, the few-shot setting with gold demonstrations substantially outperforms the zero-shot setting (78.0\% $\rightarrow$ 66.0\%). Therefore, we focus on the few-shot setting and present four pilot attempts based on two perspectives: (i) \emph{\underline{varied} / \underline{fixed}}: whether the ICL demonstrations vary for each input question; (ii) \emph{\underline{single} / \underline{mixed}}: whether the ICL demonstrations originate from a single dataset. We observe catastrophic performance degradation with these naive approaches (e.g., 78.0\% $\rightarrow$ 27.0\%). Moreover, we find that the adoption of demonstrations from a single dataset source leads to better performance as the methods with \emph{\underline{mixed}} demonstrations exhibit subpar performances than those with \emph{\underline{single}} ones (20.0/19.0\% $\rightarrow$ 26.0/27.0\%). This investigation partially inspires us to design a plug-and-play routing module to assign LLMs with demonstrations of a shared type rather than mixed types for subsequent inference.

\begin{figure*}[t]
\centering
\includegraphics[width=1.0\textwidth]{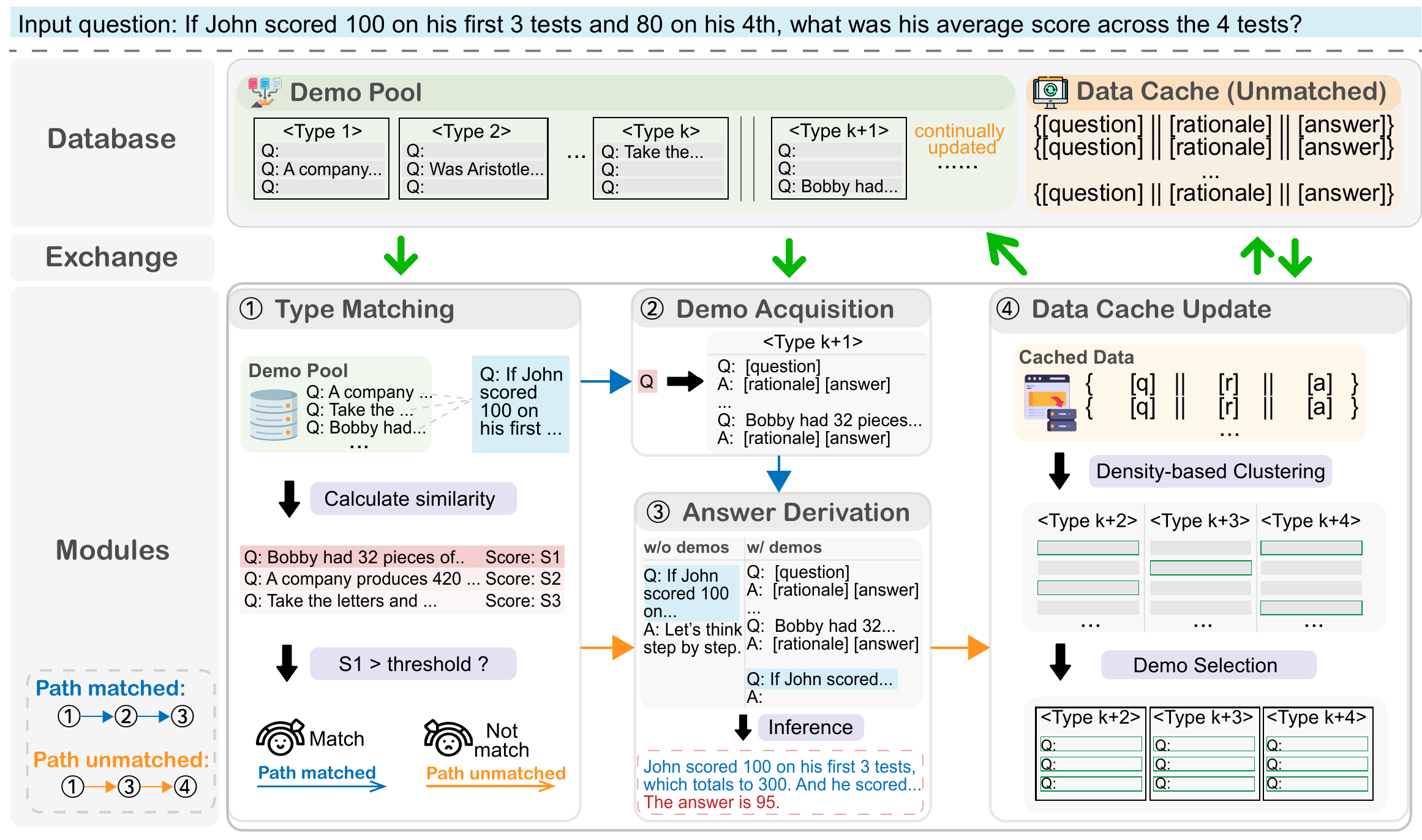}
\caption{Overview of our proposed GeM-CoT mechanism. GeM-CoT first routes the input question to different paths (\sethlcolor{greymodule}\hl{\emph{Type Matching}}): i) \textbf{\textcolor[RGB]{0,113,188}{path matched$\rightarrow$}}: For a successful match, it fetches demonstrations from the demo pool (\sethlcolor{greymodule}\hl{\emph{Demo Acquisition}}) and performs a final inference (\sethlcolor{greymodule}\hl{\emph{Answer Derivation}}). ii) \textbf{\textcolor[RGB]{255,147,30}{path unmatched$\rightarrow$}}: For a failed match, it derives the zero-shot answer with rationales (\sethlcolor{greymodule}\hl{\emph{Answer Derivation}}) and then updates the data cache through density-based clustering and automatically constructing demonstrations (\sethlcolor{greymodule}\hl{\emph{Data Cache Update}}).}
\label{fig:overview}
\end{figure*}

\begin{figure}[htb]
    \centering
    \includegraphics[width=0.43\textwidth]{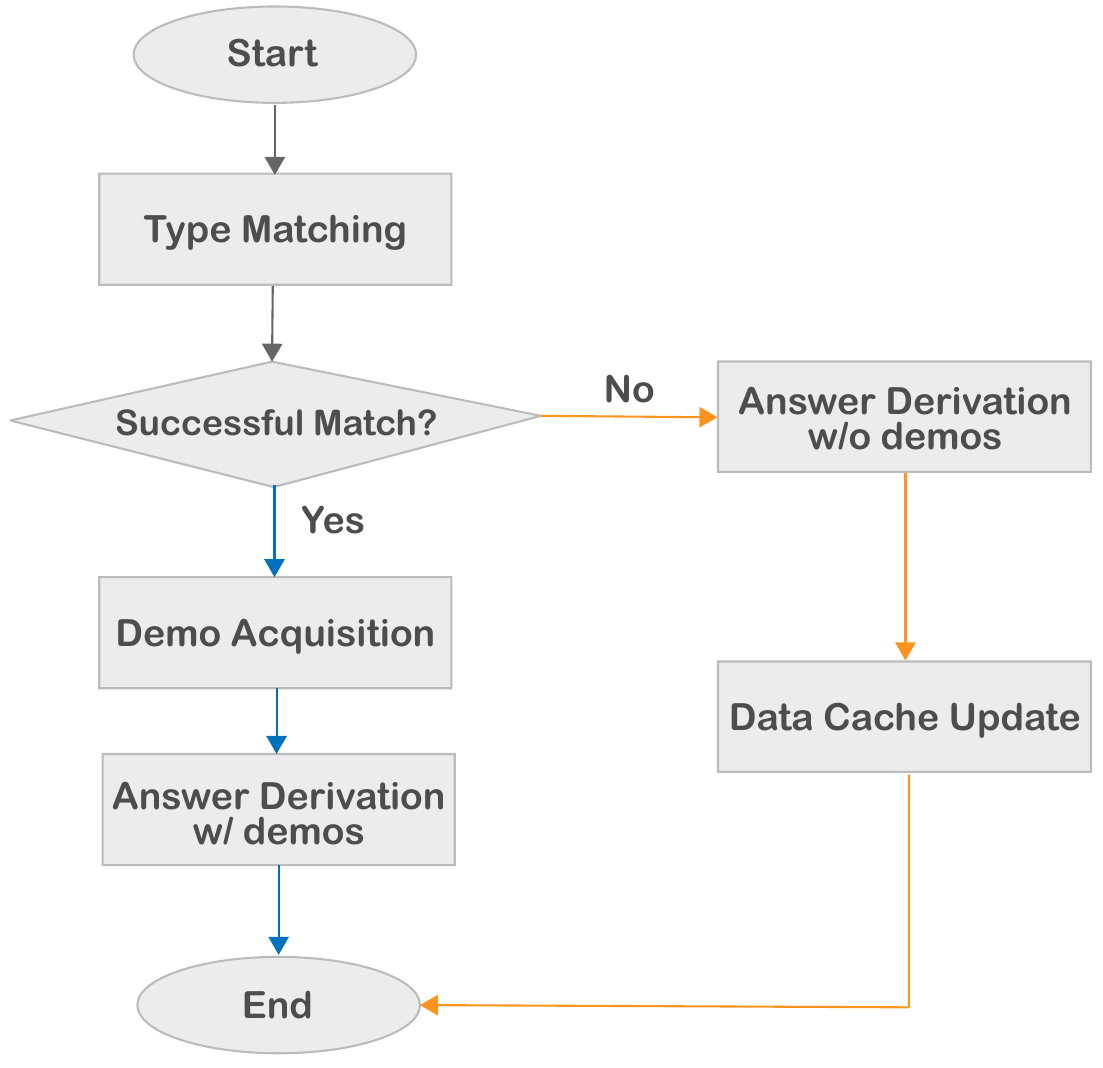}
    \caption{Flow chart of our GeM-CoT mechanism.}
    \label{fig:flow}
\end{figure}

\section{GeM-CoT}
Based on the consideration above, we introduce GeM-CoT to tackle mixed-task scenarios. 
Figure \ref{fig:overview} and Figure \ref{fig:flow} illustrate its overall architecture and flow chart, respectively.

Concretely, GeM-CoT first routes the input question to different paths (\sethlcolor{greymodule}\hl{\emph{Type Matching}}): (i) \textbf{\textcolor[RGB]{0,113,188}{path matched$\rightarrow$}}: For a successful match, it fetches demonstrations from the demo pool (\sethlcolor{greymodule}\hl{\emph{Demo Acquisition}}) and performs a final inference (\sethlcolor{greymodule}\hl{\emph{Answer Derivation}} \emph{w/ demos}). (ii) \textbf{\textcolor[RGB]{255,147,30}{path unmatched$\rightarrow$}}: For a failed match, it derives the zero-shot answer with rationales (\sethlcolor{greymodule}\hl{\emph{Answer Derivation}} \emph{w/o demos}) and then updates the data cache through density-based clustering and automatically constructs demonstrations (\sethlcolor{greymodule}\hl{\emph{Data Cache Update}}). We detail these modules as follows.

\subsection{Type Matching}\label{sec:typ_match}
Given a demo pool \texttt{DP} containing $n$ demonstrations $[{dm}^1, {dm}^2, \ldots, {dm}^n]$ and an input question $q_{in}$, the objective of \emph{Type Matching} is to find the most similar demo question for $q_{in}$ and decide whether this match is successful or not.

\paragraph{Similarity Calculation}
Note that each demonstration in \texttt{DP} is under the form: $dm^i=\left(q_d^i, r_d^i, a_d^i, t_d^i\right)$, where $r_d^i$, $a_d^i$, $t_d^i$ refer to the rationale, answer and type of $q_d^i$.
For a demo question $q_d^i \in dm^i$ and the input question $q_{in}$, we encode them independently using the same model $Enc$ and employ the dot product of their representations as the similarity score:
\begin{equation}
    sim(q_{in}, q_d^i)=\left\langle Enc(q_{in}), Enc(q_d^i)\right\rangle,
\end{equation}
where $\langle, \rangle$ denotes the dot product operation.


\paragraph{Match Decision}
After obtaining $n$ scores, we select the demonstration $dm_{sim}=\left(q_{sim}, r_{sim}, a_{sim}, t_{sim}\right)$ that has the highest similarity score with $q_{in}$: $S =sim(q_{in}, q_{sim})$. Then we compare $S$ with a constant threshold $S_{thres}$ to make a matching decision $D_{match}$:
\begin{equation}
D_{match}=\left\{\begin{array}{cl}
0, & \text { if } S \geq S_{thres}\\
1, & \text { otherwise }
\end{array}\right.
\end{equation}
For a successful match (i.e., $D_{match}=0$), we follow the path: \emph{Demo Acquisition} (\cref{sec:demo_acq}) $\rightarrow$ \emph{Answer Derivation w/ demos} (\cref{sec:ans_der}). For a failed match (i.e., $D_{match}=1$), we choose the path: \emph{Answer Derivation w/o demos} (\cref{sec:ans_der}) $\rightarrow$ \emph{Data Cache Update} (\cref{sec:data_update}).

\subsection{Demo Acquisition}\label{sec:demo_acq}
After successfully matching the input question $q_{in}$ with a certain type $t_{sim}$ in \cref{sec:typ_match}, we are able to construct type-wise demonstrations for in-context learning: $DEM_q=\left[{dm}_{q}^1, {dm}_{q}^2, \ldots, {dm}_{q}^p\right]$, where $p$ denotes the number of demonstrations under the type $t_{sim}$ in \texttt{DP}.

\subsection{Answer Derivation}\label{sec:ans_der}
\paragraph{w/ demos}
Now that we have $p$ demonstrations of the formerly matched type $t_{sim}$ acquired in \cref{sec:demo_acq}, we execute a final inference to obtain the answer to $q_{in}$. Specifically, each demonstration $dm_{q}^{i} \in DEM_{q}$ is formatted as: $\left[ \text{Q: } q^{i}, \text{A: } r^{i}, a^{i} \right]$ where $q^{i}$, $r^{i}$, and $a^{i}$ are from $dm_{q}^{i}$. Then we prepare the templated input prompt for inference by $P_{inf} = \left[ \text{Q: } q_{in}, \text{A: } \right] $. After that, the formatted demonstrations are concatenated and inserted before the input prompt $P_{inf}$, which is eventually delivered to LLMs to derive the rationale $r_{in}$ and answer $a_{in}$ of input question $q_{in}$.

\paragraph{w/o demos}
In the case of a failed match, we directly invoke Zero-Shot-CoT \citep{zero-shot} to obtain the rationale $r_{in}$ and answer $a_{in}$ for the input question $q_{in}$. Afterward, the data $\left(q_{in}, r_{in}, a_{in}\right)$ is returned to the data cache \texttt{DC}, which stores the data that undergoes a failed match with the demo pool \texttt{DP} in \emph{Type Matching} module.

\subsection{Data Cache Update}\label{sec:data_update}
Given the data cache \texttt{DC} that encompasses $m$ data $[{cad}^1, {cad}^2, \ldots, {cad}^m]$, the goal of \emph{Data Cache Update} is to execute a density-based clustering upon the questions therein and select high-quality demonstrations for each cluster that meet certain requirements. The overall procedure of this module is presented in Algorithm \ref{Algo}.

\begin{algorithm}[t]
\small
\caption{Data Cache Update}\label{Algo}
\KwIn{
    demo pool \texttt{DP}, data cache \texttt{DC},
    cached data $[{cad}^1, {cad}^2, \ldots, {cad}^m]$, 
    threshold numbers $\{th_{ca}, th_{cls}\}$, 
    density-based clustering function $\mathcal{OPTICS}$,
    demo selection function $\mathcal{SEL}$, 
    function that returns cluster size $\mathcal{S}$,
    }
\KwOut{demo pool \texttt{DP}, data cache \texttt{DC}}
\BlankLine
\If{$n \geq th_{ca}$}{
    $[{cls}^1, {cls}^2, \ldots, {cls}^t] \leftarrow \mathcal{OPTICS}([{cad}^1, {cad}^2, \ldots, {cad}^m])$ \\
    \For {$i$ in $1,...,t$}{
        $num \leftarrow \mathcal{S}({cls^i})$ \\
        \If{$num \geq th_{cls}$}{
            $demos \leftarrow \mathcal{SEL}(cls^i)$ \\
            Add $demos$ to \texttt{DP} \\
            Remove $cls^i$ from \texttt{DC}
        }
    }
}
\Return{\texttt{DP}, \texttt{DC}}
\end{algorithm}

\paragraph{Density-based Clustering}
Since the types of data in \texttt{DC} are unknown and mixed, we cannot know in advance the number of clusters into which these questions should be classified. To this end, we adopt the density-based clustering algorithm OPTICS \citep{ankerst1999optics}.\footnote{This algorithm is capable of detecting meaningful clusters in data of varied density, and this feature fits our novel setting well, where the questions are mixed and unbalanced in type.}
Concretely, we first encode all the questions $\{q_c^i \in cad^i, i \in \left[1, \ldots, m\right]\}$ in \texttt{DC} with the model $Enc$ and then perform OPTICS upon them to obtain $t$ clusters:
\begin{equation}
    \begin{split}
         \mathcal{C}_{emb} = Enc(\left[q_c^1, q_c^2, \ldots, q_c^m\right]), \\
         \left[cls^1, {cls}^2, \ldots, cls^t\right] = \text{OPTICS}(\mathcal{C}_{emb}).
    \end{split}
    \label{clus}
\end{equation}

\paragraph{Demo Selection}
After obtaining $t$ clusters, we conduct a filtering and focus only on clusters whose size is no less than a threshold $th_{cls}$. For each filtered cluster $cls^i$, we leverage the encoder model $Enc$ to obtain a vector representation for each candidate question in $cls^i$. After that, we perform $k$-means clustering over the acquired contextualized representations. We sort the questions in ascending order by distance from the cluster center. Next, we follow prior works \citep{zhang2023automatic} to conduct simple operations on the question and rationale \footnote{More details are attached in Appendix \ref{app:filtering}}, which help obtain more effective demonstrations. Once the \emph{question-rationale} pair is retained under the operation, we stop functioning on other questions in $cls^i$. As a result, we manage to collect a total of $k$ representative and high-quality demonstrations for $cls_{i}$: $[ \left(q^{1}, r^{1}, a^{1}\right), \left(q^{2}, r^{2}, a^{2}\right), \ldots, \left(q^{k}, r^{k}, a^{k}\right) ]$, where $r^{j}$ and $a^{j}$ refer to the rationale and answer of $q^{j}$. In the end, we update the demo pool \texttt{DP} with the generated diverse demonstrations and remove the data of $cls^i$ from the data cache \texttt{DC}.

\begin{table*}[t]\centering
\setlength{\tabcolsep}{1pt}
\small
\caption{Accuracy (\%) on ten reasoning datasets. The backbone model is GPT-3.5-Turbo. Results in \textbf{bold} and \underline{underline} are the best and second-best performances, respectively.}
\begin{tabular}{lcccccccccccc}
\toprule
\multirow{2}{*}{\textbf{Method}} & \textbf{Mixed-task} &\multirow{2}{*}{\textbf{AQuA}} & \multirow{2}{*}{\textbf{MultiArith}} &\multirow{2}{*}{\textbf{AddSub}} &\multirow{2}{*}{\textbf{GSM8K}} &\multirow{2}{*}{\textbf{SingleEq}} &\multirow{2}{*}{\textbf{SVAMP}} &\multirow{2}{*}{\textbf{Letter}} &\multirow{2}{*}{\textbf{Coin}} &\multirow{2}{*}{\textbf{Strategy}} &\multirow{2}{*}{\textbf{CSQA}} & \multirow{2}{*}{\textbf{Avg.}} \\
&\textbf{Scenarios} & & & & & & & & & & \\
\midrule
\midrule
\multicolumn{12}{l}{\textit{*ICL methods without CoT}}  \\
Zero-Shot &\okmark &29.1 &67.2 &88.9  &36.9  &86.5  &67.9  &4.8  &44.0  &\textbf{65.3}  &\underline{74.3} &56.5\\
Few-Shot &\ngmark  &33.1 &87.5 &91.1  &48.9  &92.7  &79.1  &7.2  &64.4  &62.3  &\textbf{81.0}  &64.7 \\
\midrule
\multicolumn{12}{l}{\textit{*Task-specific CoT approaches}}  \\
Few-Shot-CoT &\ngmark  &\textbf{54.3} &97.3 & \underline{93.9} &76.5  &{96.7}  &81.9  &73.2  &\underline{99.0}  &63.7  &78.0  &81.4\\
Auto-CoT  &\ngmark   &49.6  & \textbf{99.3} & \textbf{94.2}  &\textbf{78.9}  &96.3  &\underline{84.6}  &\underline{81.2}  &\textbf{100.0}  &\underline{64.6}  &72.2  &\underline{82.1}\\
\midrule
\multicolumn{12}{l}{\textit{*CoT techniques with \underline{generalization}}}  \\
Zero-Shot-CoT &\okmark  &51.6 &94.7 &85.5  &72.7  &93.5  &78.4  &\textbf{85.8}  &\underline{99.0}  &62.6  &69.9  &79.4 \\
General-CoT &\okmark  &46.9 &98.7 &92.4  &77.2 &\underline{97.4}  &83.8  &75.2  &\textbf{100.0}  &63.4  &72.2 &80.7 \\
\sethlcolor{mypink}\hl{\texttt{GeM-CoT}(\textbf{Ours})} &\okmark &\underline{51.9}  &\underline{99.0} & 93.7  & \underline{77.5}  &\textbf{98.4}  &\textbf{88.6}  &77.2  &\textbf{100.0}  &63.5  &72.8  
 &\textbf{82.3} \\

\bottomrule
\end{tabular}
\label{tab:results_ind}
\end{table*}

\begin{table}[t]

\centering\centering\setlength{\tabcolsep}{5pt}
\small
\caption{Accuracy (\%) on four reasoning datasets. The backbone model is GPT-4.}\label{tab:gpt4}
\begin{tabular}{lcccc}
\toprule
\textbf{Methods} & \textbf{AQuA} & \textbf{GSM8K} & \textbf{SVAMP} & \textbf{Avg.}\\

\midrule\midrule
Zero-shot-CoT &70.5 &81.3 &91.3  &81.0\\
Few-shot-CoT&71.9&92.0&90.5&85.5\\
\midrule
\sethlcolor{mypink}\hl{\texttt{GeM-CoT}(\textbf{Ours})} &\textbf{72.8}&\textbf{93.6}&\textbf{93.7}&\textbf{86.6} \\
\bottomrule
\end{tabular}
\vspace{-1mm}
\end{table}

\section{Experiments}
This section will describe our experimental setup and present the main results.
\subsection{Setup}
\paragraph{Datasets.} 
We evaluate our method on 10 reasoning datasets and a suite of 23 BIG-Bench Hard (BBH) tasks. The former is the basis of the original demo pool construction, whereas the latter can be regarded as questions of \emph{unseen}\footnote{Here \emph{unseen} means there are no questions in the original demo pool that match the BBH tasks.} types for our mechanism.
The 10 reasoning datasets include AQUA-RAT \citep{aqua}, MultiArith \citep{multiarith}, AddSub \citep{addsub}, GSM8K \citep{gsm8k}, SingleEq \citep{singleeq}, SVAMP \citep{svamp}, Last Letter Concatenation \citep{wei2023chainofthought}, Coin Flip \citep{wei2023chainofthought}, StrategyQA \citep{strategyqa}, and CSQA \citep{csqa}. For the BBH \citep{bbh} tasks, we shuffle all the data and randomly sample 2000 questions to imitate the realistic mixed-task scenarios.\footnote{Details about BBH tasks is presented in Appendix \ref{app:bbh-data-info}.}

\paragraph{Implementation.}
We utilize the popular and publicly available models GPT-3.5-Turbo and GPT-4 \citep{openai2023gpt4} from OpenAI API\footnote{\url{https://openai.com/blog/openai-api}}. The temperature and \emph{top\_p} are both set to $1.0$. 
The original demo pool \texttt{DP} is constructed based on the data from \citet{wei2023chainofthought}.
The threshold numbers $S_{thres}$, $th_{ca}$ and $th_{cls}$ are set to $0.35$, $200$ and $50$ respectively.
We employ Sentence-BERT \citep{reimers2019sentencebert} as the encoder model $Enc$.
We perform the density-based clustering and $k$-means clustering through the open-source scikit-learn\footnote{\url{https://scikit-learn.org/stable/}} python package. We set the number of demonstrations $k$ to 6 for simplicity when constructing demonstrations for a new type, since this number generally achieves decent performance on reasoning datasets \citep{wei2023chainofthought}.

\paragraph{Baselines.}
We compare GeM-CoT with 6 baselines, which can be divided into three groups: (i) ICL methods without CoT prompting \citep{zero-shot, few-shot}; (ii) task-specific CoT approaches \citep{wei2023chainofthought, zhang2023automatic}; (iii) CoT techniques with generalization \citep{zero-shot}.
Specifically, we devise a strong baseline named General-CoT for generalization comparison. It randomly collects one demonstration from each type of data in the demo pool \texttt{DP} and then leverages the gathered demonstrations as a generic inference prompt for all the input data.\footnote{The generic inference prompt is constructed from the original demo pool \texttt{DP} without subsequent updates.} More baseline details are presented in Appendix \ref{app:exp-baseline}.

\subsection{Main Results}
\paragraph{Performance on reasoning datasets.}
Table \ref{tab:results_ind} presents the results on ten reasoning tasks. 
GeM-CoT generally towers above the baseline methods from different angles. On one hand, compared with two typical task-specific CoT approaches, GeM-CoT not only averagely surpasses them in performance but also enjoys the generalizable property, which means that the input question with an unknown type can be adapted to our method in an automatic and labor-free pattern. On the other hand, while the general CoT techniques both witness average performance degradation (i.e., 82.1\%$\rightarrow$79.4/80.7\%), GeM-CoT stands out by continually boosting the performance (i.e., 82.1\%$\rightarrow$82.3\%), thus shedding light on the mutual synergy between generalization and performance.

\begin{figure}
    \centering
    \includegraphics[width=0.45\textwidth]{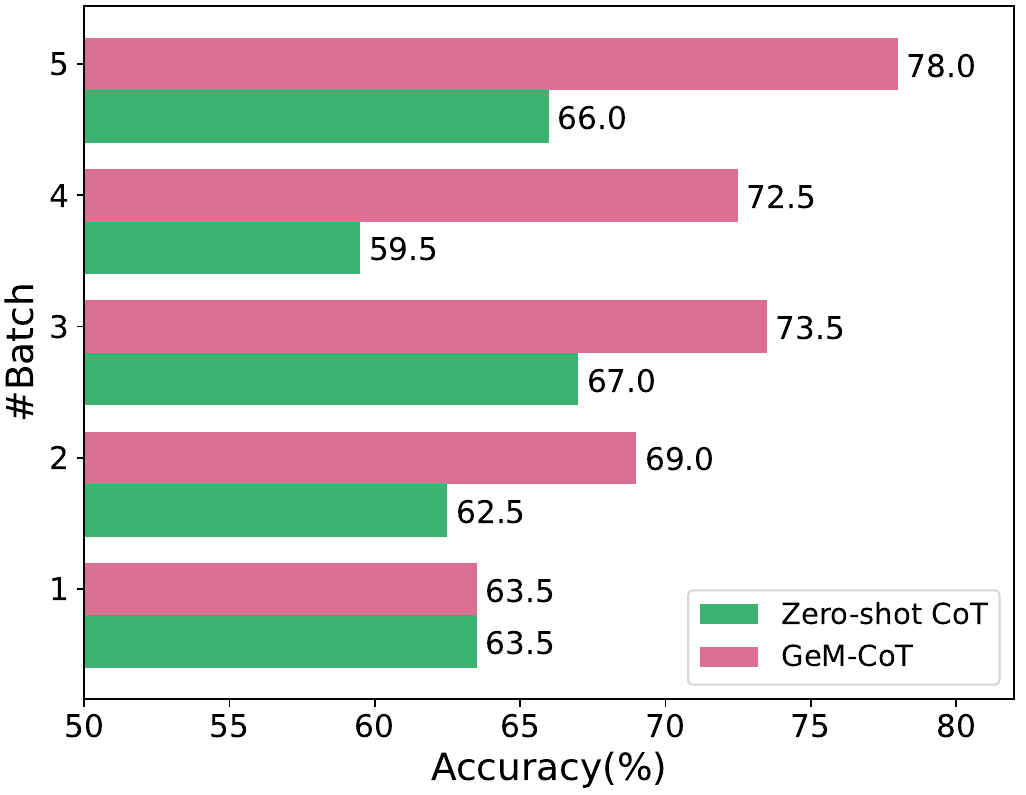}
    \caption{Process of five subsequent streaming batch data with batch size of 400 on BBH datasets.}
    \label{fig:stream}
    \vspace{-4mm}
\end{figure}

\paragraph{Performance on BBH datasets.}
As our proposed GeM-CoT is adept at tackling incoming questions of \emph{unseen} types with its continuously updating databases, we set up a more realistic and complex streaming setting \citep{tang2023chainofthought}, where the original test set is not visible and the questions appear in the form of batch data. As illustrated in Figure \ref{fig:stream}, the superiority of GeM-CoT gets prominent from batch 2, suggesting that as the data amount increases, our approach enjoys broader adaptability and higher generality by learning more representative and fine-grained features.


\section{Analysis}

\subsection{Methods of Selecting Demonstrations}
Since our work is situated in realistic mixed-task scenarios, accessing high-quality demonstrations in a labor-saving pattern is of crucial importance. Accordingly, we select two representative labor-free methods for comparison: (i) Similarity-based, which retrieves the top-$k$ similar questions based on cosine similarity; (ii) Randomness-based, which randomly samples $k$ examples for each input question. Results in Table \ref{tab:influence} show our proposed GeM-CoT (diversity-based) performs the best, verifying the importance of diversity in demonstrations.

\begin{table}
    \centering\small
        \caption{Influence of demonstration selection methods. Our proposed GeM-CoT method is based on diversity-based demonstration selection.\label{tab:influence}}
    \setlength{\tabcolsep}{5.0pt}
\begin{tabular}{lcccc}\toprule
 {Method} & {AQuA} & {AddSub} & {Strategy} & {Coin} \\
 \midrule\midrule
  \textbf{GeM-CoT} & \textbf{51.9} &\textbf{93.7} &63.5 &\textbf{100.0}\\
 \hdashline
 \quad w/ similarity &49.6 &90.1 &\textbf{64.1} &99.2\\
 \quad w/ randomness &52.0 &92.2 &61.2 &99.0 \\
 \bottomrule
\end{tabular}
\end{table}

\begin{figure}
    \centering
    \vspace{1mm}
    \includegraphics[width=0.5\textwidth]{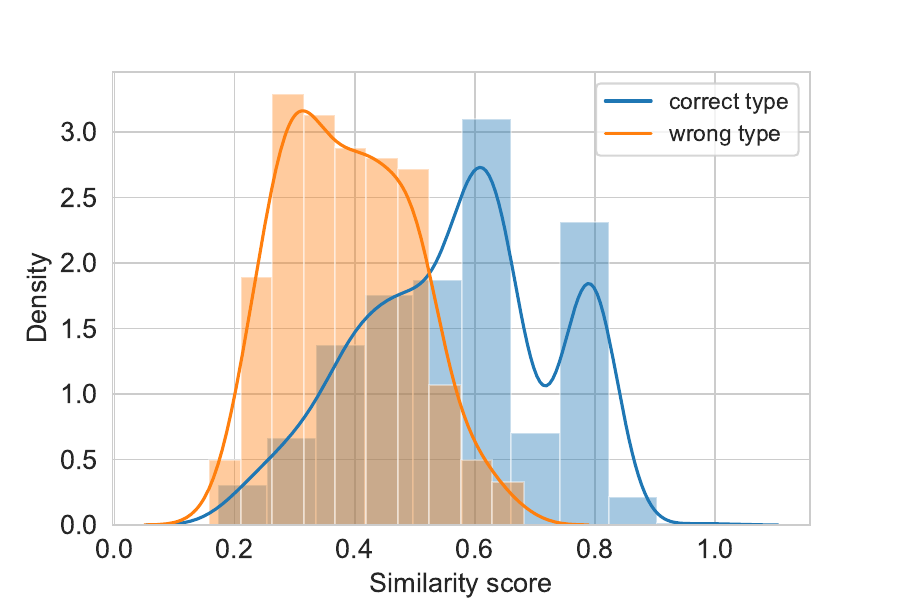}
    \caption{Distribution of similarity scores in \emph{Type Matching} module. We separately present the distribution of correctly and incorrectly matched scores.}
    \label{fig:thres_dens}
\end{figure}

\subsection{Effect of \emph{Type Matching} module}
In order to further explore the effect of \emph{Type Matching} which plays a key role in generalization, we discard this module and adopt two alternatives: (i) an LLM-based classifier that groups the questions based on its \emph{category} and \emph{form} using few-shot examples in the prompt;\footnote{We construct the few-shot examples from the ten reasoning datasets following \citep{wei2023chainofthought}. More information about how to define the \emph{category} and \emph{form} is presented in Appendix \ref{app:type}.} (ii) an idealized strategy in which we assume that the model is given the gold type, noting that this case does not apply to our proposed mixed-task scenarios, and serves only as a reference for comparison.
Results are presented in Table \ref{tab:effect}. Compared with the LLM-based classifier, GeM-CoT not only achieves comparable performance but also relieves the need for any API cost. In addition, GeM-CoT bears stronger generalization capabilities because the matching is based on semantic similarity, eliminating the effort of defining and updating the question \emph{type} in the prompt.

\begin{table}
    \centering\small
        \caption{Effect of \emph{Type Matching} module. Applicability stands for whether the method is applicable to our proposed mixed-task scenarios. \label{tab:effect}}
    \setlength{\tabcolsep}{2.1pt}
\begin{tabular}{lcccc}\toprule
 {Method}  & {{Applicability}} & {Cost-free} & {AddSub} & {Strategy}\\
 \midrule\midrule
 \textbf{GeM-CoT} &\okmark &\okmark &\textbf{93.7} &63.5 \\
  \hdashline
  \quad w/ classifier &\okmark &\ngmark &93.4 &64.5 \\
  \quad w/ correct type &\ngmark & \okmark & 90.1 & \textbf{65.0} \\
 \bottomrule
\end{tabular}
\end{table}

\subsection{Choice of Matching Threshold}
\begin{figure}
    \centering
    \vspace{1mm}
    \includegraphics[width=0.42\textwidth]{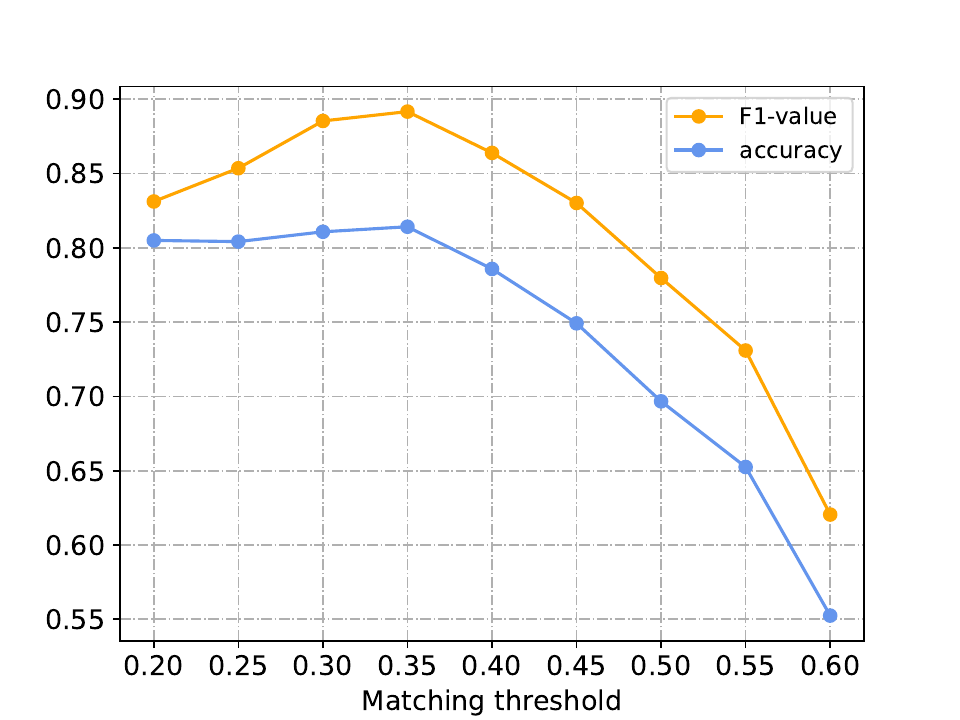}
    \caption{F1 value and accuracy of \emph{type matching} with respect to varying matching thresholds.}
    \label{fig:thres}
\end{figure}

We provide further analysis to validate the rationality of the chosen threshold for the \emph{Type Matching} module.
We focus on a total of 1200 questions from ten reasoning datasets \citep{wei2023chainofthought}, from which the original demo pool is constructed so that we can easily determine if the match types are correct or not. 
Figure \ref{fig:thres_dens} presents the distribution of correctly and incorrectly matched scores, which are concentrated in the $[0.2, 0.6]$ range. We select the scores within this range as the threshold and calculate the corresponding F1 value and accuracy. As shown in Figure \ref{fig:thres}, choosing $0.35$ yields the best results in general across our tasks.

\section{Conclusion}
In this work, we initially put forward a novel setting with significant application values, namely mixed-task scenarios where the questions come in a mixed and arbitrary way with their types unknown. Upon this challenging setting, we propose GeM-CoT, a generalizable CoT prompting mechanism that first performs type matching and then automatically samples or constructs corresponding ICL demonstrations, with continuously updated databases. Evaluation results on a total of 33 datasets demonstrate the impressive performance and superior generality of our proposed method. While most existing works focus on either promoting performance or pursuing generality, we open up a pioneering perspective to bridge the two aspects in a simple and practical manner.

\section*{Limitations}
There are two limitations. First, our proposed approach focuses on the application of CoT methods to a novel and practical scenario while ignoring the improvement of the reasoning process to a certain extent. As discussed in Related Work, existing reasoning improvement approaches can be further applied to strengthen GeM-CoT. Second, there might be more efficient ways of selecting high-quality ICL demonstrations in our proposed mixed-task scenarios, which is left to be further explored in future works.

\bibliography{anthology,custom}

\begin{thebibliography}{48}
\expandafter\ifx\csname natexlab\endcsname\relax\def\natexlab#1{#1}\fi

\bibitem[{Ankerst et~al.(1999)Ankerst, Breunig, Kriegel, and Sander}]{ankerst1999optics}
Mihael Ankerst, Markus~M Breunig, Hans-Peter Kriegel, and J{\"o}rg Sander. 1999.
\newblock Optics: Ordering points to identify the clustering structure.
\newblock \emph{ACM Sigmod record}, 28(2):49--60.

\bibitem[{Aribandi et~al.(2022)Aribandi, Tay, Schuster, Rao, Zheng, Mehta, Zhuang, Tran, Bahri, Ni, Gupta, Hui, Ruder, and Metzler}]{aribandi2022ext5}
Vamsi Aribandi, Yi~Tay, Tal Schuster, Jinfeng Rao, Huaixiu~Steven Zheng, Sanket~Vaibhav Mehta, Honglei Zhuang, Vinh~Q. Tran, Dara Bahri, Jianmo Ni, Jai~Prakash Gupta, Kai Hui, Sebastian Ruder, and Donald Metzler. 2022.
\newblock \href {https://openreview.net/forum?id=Vzh1BFUCiIX} {Ext5: Towards extreme multi-task scaling for transfer learning}.
\newblock In \emph{The Tenth International Conference on Learning Representations, {ICLR} 2022, Virtual Event, April 25-29, 2022}. OpenReview.net.

\bibitem[{Arora et~al.(2023)Arora, Narayan, Chen, Orr, Guha, Bhatia, Chami, and Re}]{arora2022ask}
Simran Arora, Avanika Narayan, Mayee~F Chen, Laurel Orr, Neel Guha, Kush Bhatia, Ines Chami, and Christopher Re. 2023.
\newblock Ask me anything: A simple strategy for prompting language models.
\newblock In \emph{The Eleventh International Conference on Learning Representations}.

\bibitem[{Brown et~al.(2020)Brown, Mann, Ryder, Subbiah, Kaplan, Dhariwal, Neelakantan, Shyam, Sastry, Askell, Agarwal, Herbert{-}Voss, Krueger, Henighan, Child, Ramesh, Ziegler, Wu, Winter, Hesse, Chen, Sigler, Litwin, Gray, Chess, Clark, Berner, McCandlish, Radford, Sutskever, and Amodei}]{few-shot}
Tom~B. Brown, Benjamin Mann, Nick Ryder, Melanie Subbiah, Jared Kaplan, Prafulla Dhariwal, Arvind Neelakantan, Pranav Shyam, Girish Sastry, Amanda Askell, Sandhini Agarwal, Ariel Herbert{-}Voss, Gretchen Krueger, Tom Henighan, Rewon Child, Aditya Ramesh, Daniel~M. Ziegler, Jeffrey Wu, Clemens Winter, Christopher Hesse, Mark Chen, Eric Sigler, Mateusz Litwin, Scott Gray, Benjamin Chess, Jack Clark, Christopher Berner, Sam McCandlish, Alec Radford, Ilya Sutskever, and Dario Amodei. 2020.
\newblock \href {https://proceedings.neurips.cc/paper/2020/hash/1457c0d6bfcb4967418bfb8ac142f64a-Abstract.html} {Language models are few-shot learners}.
\newblock In \emph{Advances in Neural Information Processing Systems 33: Annual Conference on Neural Information Processing Systems 2020, NeurIPS 2020, December 6-12, 2020, virtual}.

\bibitem[{Chowdhery et~al.(2022)Chowdhery, Narang, Devlin, Bosma, Mishra, Roberts, Barham, Chung, Sutton, Gehrmann et~al.}]{chowdhery2022palm}
Aakanksha Chowdhery, Sharan Narang, Jacob Devlin, Maarten Bosma, Gaurav Mishra, Adam Roberts, Paul Barham, Hyung~Won Chung, Charles Sutton, Sebastian Gehrmann, et~al. 2022.
\newblock \href {https://arxiv.org/abs/2204.02311} {Palm: Scaling language modeling with pathways}.
\newblock \emph{ArXiv preprint}, abs/2204.02311.

\bibitem[{Cobbe et~al.(2021)Cobbe, Kosaraju, Bavarian, Chen, Jun, Kaiser, Plappert, Tworek, Hilton, Nakano et~al.}]{gsm8k}
Karl Cobbe, Vineet Kosaraju, Mohammad Bavarian, Mark Chen, Heewoo Jun, Lukasz Kaiser, Matthias Plappert, Jerry Tworek, Jacob Hilton, Reiichiro Nakano, et~al. 2021.
\newblock \href {https://arxiv.org/abs/2110.14168} {Training verifiers to solve math word problems}.
\newblock \emph{ArXiv preprint}, abs/2110.14168.

\bibitem[{Creswell et~al.(2023)Creswell, Shanahan, and Higgins}]{creswell2022selectioninference}
Antonia Creswell, Murray Shanahan, and Irina Higgins. 2023.
\newblock Selection-inference: Exploiting large language models for interpretable logical reasoning.
\newblock In \emph{The Eleventh International Conference on Learning Representations}.

\bibitem[{Diao et~al.(2023)Diao, Wang, Lin, and Zhang}]{diao2023active}
Shizhe Diao, Pengcheng Wang, Yong Lin, and Tong Zhang. 2023.
\newblock \href {https://arxiv.org/abs/2302.12246} {Active prompting with chain-of-thought for large language models}.
\newblock \emph{ArXiv preprint}, abs/2302.12246.

\bibitem[{Geva et~al.(2021)Geva, Khashabi, Segal, Khot, Roth, and Berant}]{strategyqa}
Mor Geva, Daniel Khashabi, Elad Segal, Tushar Khot, Dan Roth, and Jonathan Berant. 2021.
\newblock \href {https://doi.org/10.1162/tacl_a_00370} {Did aristotle use a laptop? a question answering benchmark with implicit reasoning strategies}.
\newblock \emph{Transactions of the Association for Computational Linguistics}, 9:346--361.

\bibitem[{Ho et~al.(2022)Ho, Schmid, and Yun}]{ho2022large}
Namgyu Ho, Laura Schmid, and Se-Young Yun. 2022.
\newblock \href {https://arxiv.org/abs/2212.10071} {Large language models are reasoning teachers}.
\newblock \emph{ArXiv preprint}, abs/2212.10071.

\bibitem[{Hosseini et~al.(2014)Hosseini, Hajishirzi, Etzioni, and Kushman}]{addsub}
Mohammad~Javad Hosseini, Hannaneh Hajishirzi, Oren Etzioni, and Nate Kushman. 2014.
\newblock \href {https://doi.org/10.3115/v1/D14-1058} {Learning to solve arithmetic word problems with verb categorization}.
\newblock In \emph{Proceedings of the 2014 Conference on Empirical Methods in Natural Language Processing ({EMNLP})}, pages 523--533, Doha, Qatar. Association for Computational Linguistics.

\bibitem[{Huang et~al.(2023)Huang, Liu, Lin, Pang, Du, and Lin}]{huang2023lorahub}
Chengsong Huang, Qian Liu, Bill~Yuchen Lin, Tianyu Pang, Chao Du, and Min Lin. 2023.
\newblock \href {https://arxiv.org/abs/2307.13269} {Lorahub: Efficient cross-task generalization via dynamic lora composition}.
\newblock \emph{ArXiv preprint}, abs/2307.13269.

\bibitem[{Kojima et~al.(2023)Kojima, Gu, Reid, Matsuo, and Iwasawa}]{zero-shot}
Takeshi Kojima, Shixiang~(Shane) Gu, Machel Reid, Yutaka Matsuo, and Yusuke Iwasawa. 2023.
\newblock Large language models are zero-shot reasoners.
\newblock In \emph{Advances in Neural Information Processing Systems}, volume~35, pages 22199--22213.

\bibitem[{Koncel-Kedziorski et~al.(2015)Koncel-Kedziorski, Hajishirzi, Sabharwal, Etzioni, and Ang}]{singleeq}
Rik Koncel-Kedziorski, Hannaneh Hajishirzi, Ashish Sabharwal, Oren Etzioni, and Siena~Dumas Ang. 2015.
\newblock \href {https://doi.org/10.1162/tacl_a_00160} {Parsing algebraic word problems into equations}.
\newblock \emph{Transactions of the Association for Computational Linguistics}, 3:585--597.

\bibitem[{Koncel-Kedziorski et~al.(2016)Koncel-Kedziorski, Roy, Amini, Kushman, and Hajishirzi}]{koncel-kedziorski-etal-2016-mawps}
Rik Koncel-Kedziorski, Subhro Roy, Aida Amini, Nate Kushman, and Hannaneh Hajishirzi. 2016.
\newblock \href {https://doi.org/10.18653/v1/N16-1136} {{MAWPS}: A math word problem repository}.
\newblock In \emph{Proceedings of the 2016 Conference of the North {A}merican Chapter of the Association for Computational Linguistics: Human Language Technologies}, pages 1152--1157, San Diego, California. Association for Computational Linguistics.

\bibitem[{Li et~al.(2023)Li, Lin, Zhang, Fu, Chen, Lou, and Chen}]{li2023making}
Yifei Li, Zeqi Lin, Shizhuo Zhang, Qiang Fu, Bei Chen, Jian-Guang Lou, and Weizhu Chen. 2023.
\newblock Making language models better reasoners with step-aware verifier.
\newblock In \emph{Proceedings of the 61st Annual Meeting of the Association for Computational Linguistics (Volume 1: Long Papers)}, pages 5315--5333.

\bibitem[{Ling et~al.(2017)Ling, Yogatama, Dyer, and Blunsom}]{aqua}
Wang Ling, Dani Yogatama, Chris Dyer, and Phil Blunsom. 2017.
\newblock \href {https://doi.org/10.18653/v1/P17-1015} {Program induction by rationale generation: Learning to solve and explain algebraic word problems}.
\newblock In \emph{Proceedings of the 55th Annual Meeting of the Association for Computational Linguistics (Volume 1: Long Papers)}, pages 158--167, Vancouver, Canada. Association for Computational Linguistics.

\bibitem[{Liu et~al.(2022)Liu, Shen, Zhang, Dolan, Carin, and Chen}]{liu-etal-2022-makes}
Jiachang Liu, Dinghan Shen, Yizhe Zhang, Bill Dolan, Lawrence Carin, and Weizhu Chen. 2022.
\newblock \href {https://doi.org/10.18653/v1/2022.deelio-1.10} {What makes good in-context examples for {GPT}-3?}
\newblock In \emph{Proceedings of Deep Learning Inside Out (DeeLIO 2022): The 3rd Workshop on Knowledge Extraction and Integration for Deep Learning Architectures}, pages 100--114, Dublin, Ireland and Online. Association for Computational Linguistics.

\bibitem[{Liu et~al.(2019)Liu, He, Chen, and Gao}]{liu2019multitask}
Xiaodong Liu, Pengcheng He, Weizhu Chen, and Jianfeng Gao. 2019.
\newblock \href {https://doi.org/10.18653/v1/P19-1441} {Multi-task deep neural networks for natural language understanding}.
\newblock In \emph{Proceedings of the 57th Annual Meeting of the Association for Computational Linguistics}, pages 4487--4496, Florence, Italy. Association for Computational Linguistics.

\bibitem[{Lyu et~al.(2023)Lyu, Havaldar, Stein, Zhang, Rao, Wong, Apidianaki, and Callison-Burch}]{lyu2023faithful}
Qing Lyu, Shreya Havaldar, Adam Stein, Li~Zhang, Delip Rao, Eric Wong, Marianna Apidianaki, and Chris Callison-Burch. 2023.
\newblock \href {https://arxiv.org/abs/2301.13379} {Faithful chain-of-thought reasoning}.
\newblock \emph{ArXiv preprint}, abs/2301.13379.

\bibitem[{Madaan and Yazdanbakhsh(2022)}]{madaan2022text}
Aman Madaan and Amir Yazdanbakhsh. 2022.
\newblock \href {https://arxiv.org/abs/2209.07686} {Text and patterns: For effective chain of thought, it takes two to tango}.
\newblock \emph{ArXiv preprint}, abs/2209.07686.

\bibitem[{OpenAI(2023)}]{openai2023gpt4}
OpenAI. 2023.
\newblock \href {https://arxiv.org/abs/2303.08774} {Gpt-4 technical report}.
\newblock \emph{ArXiv preprint}, abs/2303.08774.

\bibitem[{Patel et~al.(2021)Patel, Bhattamishra, and Goyal}]{svamp}
Arkil Patel, Satwik Bhattamishra, and Navin Goyal. 2021.
\newblock \href {https://doi.org/10.18653/v1/2021.naacl-main.168} {Are {NLP} models really able to solve simple math word problems?}
\newblock In \emph{Proceedings of the 2021 Conference of the North American Chapter of the Association for Computational Linguistics: Human Language Technologies}, pages 2080--2094, Online. Association for Computational Linguistics.

\bibitem[{Press et~al.(2022)Press, Zhang, Min, Schmidt, Smith, and Lewis}]{press2022measuring}
Ofir Press, Muru Zhang, Sewon Min, Ludwig Schmidt, Noah~A Smith, and Mike Lewis. 2022.
\newblock \href {https://arxiv.org/abs/2210.03350} {Measuring and narrowing the compositionality gap in language models}.
\newblock \emph{ArXiv preprint}, abs/2210.03350.

\bibitem[{Reimers and Gurevych(2019)}]{reimers2019sentencebert}
Nils Reimers and Iryna Gurevych. 2019.
\newblock \href {https://doi.org/10.18653/v1/D19-1410} {Sentence-{BERT}: Sentence embeddings using {S}iamese {BERT}-networks}.
\newblock In \emph{Proceedings of the 2019 Conference on Empirical Methods in Natural Language Processing and the 9th International Joint Conference on Natural Language Processing (EMNLP-IJCNLP)}, pages 3982--3992, Hong Kong, China. Association for Computational Linguistics.

\bibitem[{Roy and Roth(2015)}]{multiarith}
Subhro Roy and Dan Roth. 2015.
\newblock \href {https://doi.org/10.18653/v1/D15-1202} {Solving general arithmetic word problems}.
\newblock In \emph{Proceedings of the 2015 Conference on Empirical Methods in Natural Language Processing}, pages 1743--1752, Lisbon, Portugal. Association for Computational Linguistics.

\bibitem[{Sanh et~al.(2022)Sanh, Webson, Raffel, Bach, Sutawika, Alyafeai, Chaffin, Stiegler, Raja, Dey, Bari, Xu, Thakker, Sharma, Szczechla, Kim, Chhablani, Nayak, Datta, Chang, Jiang, Wang, Manica, Shen, Yong, Pandey, Bawden, Wang, Neeraj, Rozen, Sharma, Santilli, F{\'{e}}vry, Fries, Teehan, Scao, Biderman, Gao, Wolf, and Rush}]{sanh2022multitask}
Victor Sanh, Albert Webson, Colin Raffel, Stephen~H. Bach, Lintang Sutawika, Zaid Alyafeai, Antoine Chaffin, Arnaud Stiegler, Arun Raja, Manan Dey, M~Saiful Bari, Canwen Xu, Urmish Thakker, Shanya~Sharma Sharma, Eliza Szczechla, Taewoon Kim, Gunjan Chhablani, Nihal~V. Nayak, Debajyoti Datta, Jonathan Chang, Mike~Tian{-}Jian Jiang, Han Wang, Matteo Manica, Sheng Shen, Zheng~Xin Yong, Harshit Pandey, Rachel Bawden, Thomas Wang, Trishala Neeraj, Jos Rozen, Abheesht Sharma, Andrea Santilli, Thibault F{\'{e}}vry, Jason~Alan Fries, Ryan Teehan, Teven~Le Scao, Stella Biderman, Leo Gao, Thomas Wolf, and Alexander~M. Rush. 2022.
\newblock \href {https://openreview.net/forum?id=9Vrb9D0WI4} {Multitask prompted training enables zero-shot task generalization}.
\newblock In \emph{The Tenth International Conference on Learning Representations, {ICLR} 2022, Virtual Event, April 25-29, 2022}. OpenReview.net.

\bibitem[{Scao et~al.(2022)Scao, Fan, Akiki, Pavlick, Ili{\'c}, Hesslow, Castagn{\'e}, Luccioni, Yvon, Gall{\'e} et~al.}]{scao2022bloom}
Teven~Le Scao, Angela Fan, Christopher Akiki, Ellie Pavlick, Suzana Ili{\'c}, Daniel Hesslow, Roman Castagn{\'e}, Alexandra~Sasha Luccioni, Fran{\c{c}}ois Yvon, Matthias Gall{\'e}, et~al. 2022.
\newblock \href {https://arxiv.org/abs/2211.05100} {Bloom: A 176b-parameter open-access multilingual language model}.
\newblock \emph{ArXiv preprint}, abs/2211.05100.

\bibitem[{Suzgun et~al.(2022)Suzgun, Scales, Sch{\"a}rli, Gehrmann, Tay, Chung, Chowdhery, Le, Chi, Zhou et~al.}]{bbh}
Mirac Suzgun, Nathan Scales, Nathanael Sch{\"a}rli, Sebastian Gehrmann, Yi~Tay, Hyung~Won Chung, Aakanksha Chowdhery, Quoc~V Le, Ed~H Chi, Denny Zhou, et~al. 2022.
\newblock \href {https://arxiv.org/abs/2210.09261} {Challenging big-bench tasks and whether chain-of-thought can solve them}.
\newblock \emph{ArXiv preprint}, abs/2210.09261.

\bibitem[{Talmor et~al.(2019)Talmor, Herzig, Lourie, and Berant}]{csqa}
Alon Talmor, Jonathan Herzig, Nicholas Lourie, and Jonathan Berant. 2019.
\newblock \href {https://doi.org/10.18653/v1/N19-1421} {{C}ommonsense{QA}: A question answering challenge targeting commonsense knowledge}.
\newblock In \emph{Proceedings of the 2019 Conference of the North {A}merican Chapter of the Association for Computational Linguistics: Human Language Technologies, Volume 1 (Long and Short Papers)}, pages 4149--4158, Minneapolis, Minnesota. Association for Computational Linguistics.

\bibitem[{Tang(2023)}]{tang2023chainofthought}
Yuxin Tang. 2023.
\newblock \href {http://arxiv.org/abs/2306.00550} {Chain-of-thought prompting under streaming batch: A case study}.
\newblock \emph{arXiv}.

\bibitem[{Thoppilan et~al.(2022)Thoppilan, De~Freitas, Hall, Shazeer, Kulshreshtha, Cheng, Jin, Bos, Baker, Du et~al.}]{thoppilan2022lamda}
Romal Thoppilan, Daniel De~Freitas, Jamie Hall, Noam Shazeer, Apoorv Kulshreshtha, Heng-Tze Cheng, Alicia Jin, Taylor Bos, Leslie Baker, Yu~Du, et~al. 2022.
\newblock \href {https://arxiv.org/abs/2201.08239} {Lamda: Language models for dialog applications}.
\newblock \emph{ArXiv preprint}, abs/2201.08239.

\bibitem[{Touvron et~al.(2023)Touvron, Lavril, Izacard, Martinet, Lachaux, Lacroix, Rozi{\`e}re, Goyal, Hambro, Azhar et~al.}]{touvron2023llama}
Hugo Touvron, Thibaut Lavril, Gautier Izacard, Xavier Martinet, Marie-Anne Lachaux, Timoth{\'e}e Lacroix, Baptiste Rozi{\`e}re, Naman Goyal, Eric Hambro, Faisal Azhar, et~al. 2023.
\newblock \href {https://arxiv.org/abs/2302.13971} {Llama: Open and efficient foundation language models}.
\newblock \emph{ArXiv preprint}, abs/2302.13971.

\bibitem[{Wan et~al.(2023)Wan, Sun, Dai, Arik, and Pfister}]{wan-etal-2023-better}
Xingchen Wan, Ruoxi Sun, Hanjun Dai, Sercan Arik, and Tomas Pfister. 2023.
\newblock \href {https://doi.org/10.18653/v1/2023.findings-acl.216} {Better zero-shot reasoning with self-adaptive prompting}.
\newblock In \emph{Findings of the Association for Computational Linguistics: ACL 2023}, pages 3493--3514, Toronto, Canada. Association for Computational Linguistics.

\bibitem[{Wang et~al.(2022)Wang, Wei, Schuurmans, Le, Chi, and Zhou}]{wang2022rationaleaugmented}
Xuezhi Wang, Jason Wei, Dale Schuurmans, Quoc Le, Ed~Chi, and Denny Zhou. 2022.
\newblock \href {https://arxiv.org/abs/2207.00747} {Rationale-augmented ensembles in language models}.
\newblock \emph{ArXiv preprint}, abs/2207.00747.

\bibitem[{Wang et~al.(2023{\natexlab{a}})Wang, Wei, Schuurmans, Le, Chi, Narang, Chowdhery, and Zhou}]{wang2023selfconsistency}
Xuezhi Wang, Jason Wei, Dale Schuurmans, Quoc~V Le, Ed~H Chi, Sharan Narang, Aakanksha Chowdhery, and Denny Zhou. 2023{\natexlab{a}}.
\newblock Self-consistency improves chain of thought reasoning in language models.
\newblock In \emph{The Eleventh International Conference on Learning Representations}.

\bibitem[{Wang et~al.(2023{\natexlab{b}})Wang, Zhang, and Wang}]{wang2023meta}
Yiming Wang, Zhuosheng Zhang, and Rui Wang. 2023{\natexlab{b}}.
\newblock Meta-reasoning: Semantics-symbol deconstruction for large language models.
\newblock \emph{arXiv preprint arXiv:2306.17820}.

\bibitem[{Wei et~al.(2022)Wei, Bosma, Zhao, Guu, Yu, Lester, Du, Dai, and Le}]{wei2022finetuned}
Jason Wei, Maarten Bosma, Vincent~Y. Zhao, Kelvin Guu, Adams~Wei Yu, Brian Lester, Nan Du, Andrew~M. Dai, and Quoc~V. Le. 2022.
\newblock \href {https://openreview.net/forum?id=gEZrGCozdqR} {Finetuned language models are zero-shot learners}.
\newblock In \emph{The Tenth International Conference on Learning Representations, {ICLR} 2022, Virtual Event, April 25-29, 2022}. OpenReview.net.

\bibitem[{Wei et~al.(2023)Wei, Wang, Schuurmans, Bosma, Xia, Chi, Le, Zhou et~al.}]{wei2023chainofthought}
Jason Wei, Xuezhi Wang, Dale Schuurmans, Maarten Bosma, Fei Xia, Ed~Chi, Quoc~V Le, Denny Zhou, et~al. 2023.
\newblock Chain-of-thought prompting elicits reasoning in large language models.
\newblock \emph{Advances in Neural Information Processing Systems}, 35:24824--24837.

\bibitem[{Weng et~al.(2022)Weng, Zhu, Xia, Li, He, Liu, and Zhao}]{weng2023large}
Yixuan Weng, Minjun Zhu, Fei Xia, Bin Li, Shizhu He, Kang Liu, and Jun Zhao. 2022.
\newblock \href {https://arxiv.org/abs/2212.09561} {Large language models are better reasoners with self-verification}.
\newblock \emph{ArXiv preprint}, abs/2212.09561.

\bibitem[{Yang et~al.(2023)Yang, Wang, Lu, Liu, Le, Zhou, and Chen}]{yang2023large}
Chengrun Yang, Xuezhi Wang, Yifeng Lu, Hanxiao Liu, Quoc~V Le, Denny Zhou, and Xinyun Chen. 2023.
\newblock \href {https://arxiv.org/abs/2309.03409} {Large language models as optimizers}.
\newblock \emph{ArXiv preprint}, abs/2309.03409.

\bibitem[{Yoran et~al.(2023)Yoran, Wolfson, Bogin, Katz, Deutch, and Berant}]{yoran2023answering}
Ori Yoran, Tomer Wolfson, Ben Bogin, Uri Katz, Daniel Deutch, and Jonathan Berant. 2023.
\newblock \href {https://arxiv.org/abs/2304.13007} {Answering questions by meta-reasoning over multiple chains of thought}.
\newblock \emph{ArXiv preprint}, abs/2304.13007.

\bibitem[{Yu et~al.(2022)Yu, Zhu, Fang, Yu, Wang, Xu, Zeng, and Jiang}]{yu2022dictbert}
Wenhao Yu, Chenguang Zhu, Yuwei Fang, Donghan Yu, Shuohang Wang, Yichong Xu, Michael Zeng, and Meng Jiang. 2022.
\newblock \href {https://doi.org/10.18653/v1/2022.findings-acl.150} {Dict-{BERT}: Enhancing language model pre-training with dictionary}.
\newblock In \emph{Findings of the Association for Computational Linguistics: ACL 2022}, pages 1907--1918, Dublin, Ireland. Association for Computational Linguistics.

\bibitem[{Zhang et~al.(2022)Zhang, Wang, Xu, Fang, Yu, Liu, Zhao, Zhu, and Zeng}]{zhang-etal-2022-task}
Zhuosheng Zhang, Shuohang Wang, Yichong Xu, Yuwei Fang, Wenhao Yu, Yang Liu, Hai Zhao, Chenguang Zhu, and Michael Zeng. 2022.
\newblock \href {https://aclanthology.org/2022.findings-emnlp.416} {Task compass: Scaling multi-task pre-training with task prefix}.
\newblock In \emph{Findings of the Association for Computational Linguistics: EMNLP 2022}, pages 5671--5685, Abu Dhabi, United Arab Emirates. Association for Computational Linguistics.

\bibitem[{Zhang et~al.(2023)Zhang, Zhang, Li, and Smola}]{zhang2023automatic}
Zhuosheng Zhang, Aston Zhang, Mu~Li, and Alex Smola. 2023.
\newblock Automatic chain of thought prompting in large language models.
\newblock In \emph{The Eleventh International Conference on Learning Representations (ICLR 2023)}.

\bibitem[{Zhang and Zhao(2021)}]{zhang-zhao-2021-structural}
Zhuosheng Zhang and Hai Zhao. 2021.
\newblock \href {https://doi.org/10.18653/v1/2021.acl-long.399} {Structural pre-training for dialogue comprehension}.
\newblock In \emph{Proceedings of the 59th Annual Meeting of the Association for Computational Linguistics and the 11th International Joint Conference on Natural Language Processing (Volume 1: Long Papers)}, pages 5134--5145, Online. Association for Computational Linguistics.

\bibitem[{Zhou et~al.(2023)Zhou, Sch{\"a}rli, Hou, Wei, Scales, Wang, Schuurmans, Cui, Bousquet, Le et~al.}]{zhou2023leasttomost}
Denny Zhou, Nathanael Sch{\"a}rli, Le~Hou, Jason Wei, Nathan Scales, Xuezhi Wang, Dale Schuurmans, Claire Cui, Olivier Bousquet, Quoc~V Le, et~al. 2023.
\newblock Least-to-most prompting enables complex reasoning in large language models.
\newblock In \emph{The Eleventh International Conference on Learning Representations}.

\bibitem[{Zhou et~al.(2022)Zhou, Muresanu, Han, Paster, Pitis, Chan, and Ba}]{zhou2022large}
Yongchao Zhou, Andrei~Ioan Muresanu, Ziwen Han, Keiran Paster, Silviu Pitis, Harris Chan, and Jimmy Ba. 2022.
\newblock Large language models are human-level prompt engineers.
\newblock In \emph{The Eleventh International Conference on Learning Representations}.

\end{thebibliography}

\appendix

\section{Experimental Details}\label{app:exp}
\subsection{Implementation details}\label{app:filtering}
\paragraph{Filtering operations in \emph{Demo Selection}.}
We follow the works from \citep{wei2023chainofthought,zhang2023automatic} to filter the \emph{question-rationale} pair as follows: the question needs to be no more than 60 tokens and the rationale should not exceed 5 reasoning steps. 
The objective of this filtering strategy is to seek simple heuristics by sampling simpler questions and rationales. 

\subsection{Baseline Methods}\label{app:exp-baseline}
We introduce the baseline methods in detail.

$\bullet$ \quad \textbf{ICL methods without CoT}: Zero-Shot \citep{zero-shot} adds the prompt ``A: The answer is'' to an input question and leverage it as the input delivered to LLMs. Few-Shot \citep{few-shot} employs several additional templated demonstrations as: $\left[ \text{Q: } \texttt{q}, \text{A: The answer is } \texttt{a}  \right]$ before the input question, where \texttt{q} and \texttt{a} are manually crafted questions and answers.

$\bullet$ \quad \textbf{Task-specific CoT approaches.}: Few-Shot-CoT \citep{wei2023chainofthought} follows similar patterns as Few-Shot but differs in that rationales are inserted before deriving the answer. Auto-CoT \citep{zhang2023automatic} divides questions of a given dataset into a few clusters, samples a representative question from each cluster, and constructs its reasoning chain using Zero-Shot-CoT with simple heuristics.

$\bullet$ \quad \textbf{CoT techniques with generalization}: Zero-Shot-CoT \citep{zero-shot} simply inserts the prompt \emph{Let's think step by step} after a question to conduct inference, which rids the necessity of handcrafted task-wise demonstrations. We also compare our method with a strong baseline General-CoT, in which the in-context demonstrations for inference come from distinct question groups.

\section{Dataset Information}\label{app:data-info}
\subsection{Reasoning Datasets}\label{app:reason-data-info}
Our method is evaluated on 10 reasoning benchmark datasets that cover three categories including arithmetic, commonsense and symbolic tasks and involve three forms encompassing short-answer, multiple-choice, and yes-or-no questions. The corresponding categories and forms of these datasets are shown in Table \ref{tab:indomain-data}.

$\bullet$ \quad \textbf{Arithmetic Reasoning}: we choose the following six datasets: (i) MultiArith \citep{multiarith}, (ii) GSM8K \citep{gsm8k}, (iii) AddSub \citep{addsub}, (iv) AQUA-RAT \citep{aqua}, (v) SingleEq \citep{singleeq}, and (vi) SVAMP \citep{svamp}. MultiArith, AddSub, and SingleEq come from the Math World Problem Repository \citep{koncel-kedziorski-etal-2016-mawps}, while the other three are from more contemporary benchmarks. Among them, all the arithmetic datasets belong to short-answer form except for AQUA-RAT which is in multiple-choice format.

$\bullet$ \quad \textbf{Commonsense Reasoning}: we take the following two datasets into account: (i) CSQA \citep{csqa} and StrategyQA \citep{strategyqa}. CSQA poses difficult questions with rich semantic relations by making use of ConceptNet \citep{csqa}. StrategyQA requires models to derive answers using implicit reasoning steps \citep{strategyqa}. CSQA is in multiple-choice form whereas StrategyQA belongs to the yes-or-no format.

$\bullet$ \quad \textbf{Symbolic Reasoning}: we employ the typical datasets Last Letter Concatenation and Coin Flip from \citet{wei2023chainofthought}, which are in short-answer and yes-or-no form respectively. Last Letter Concatenation asks the model to concatenate the last letters of each word. Coin Filp requires the model to answer whether a coin heads up after a series of actions of either flipping or not flipping the coin.

\begin{table*}[htb]\centering
\small
\setlength{\tabcolsep}{7pt}
\caption{Information of 10 reasoning datasets (Ari.: arithmetic; Com.: commonsense and Sym.: symbolic; SAQ: short-answer question; MCQ: multiple-choice question; Y/N: yes-or-no question).
}
\vspace{2.8mm}
\begin{tabular}{lcccccccccc}
\toprule
Task &MultiArith &GSM8K &AddSub &AQuA &SingleEq &SVAMP &CSQA &Strategy &Letter &Coin \\
\midrule
Category  &Ari. &Ari. &Ari. &Ari. &Ari. &Ari. &Com. &Com. &Sym. &Sym.\\
Form  &SAQ  &SAQ &SAQ &MCQ &SAQ &SAQ &MCQ &Y/N &SAQ &Y/N\\
Size &600  &1319 &395 &254 &508 &1000 &1221 &2290 &500 &500\\
\bottomrule
\end{tabular}
\vspace{-3.6mm}
\label{tab:indomain-data}
\end{table*}

\subsection{BBH Datasets}\label{app:bbh-data-info}
We further evaluate our method on a suite of 23 BBH tasks, the questions of which can be regarded as \emph{unseen} types for our proposed mechanism. The detailed information about these BBH datasets are listed in Table \ref{tab:bbh-data}.

\begin{table*}[htb]\centering
\small
\setlength{\tabcolsep}{3pt}
\caption{Information of 23 BBH datasets. Categories and descriptions about the datasets are from \citet{bbh}. (Algo.+Ari.: Algorithmic and Multi-Step Arithmetic Reasoning; NLU: Natural Language Understanding; Knowledge: Use of World Knowledge).}
\begin{tabular}{p{0.20\linewidth}p{0.18\linewidth}p{0.62\linewidth}}
\toprule
Task & Category & Description \\
\midrule
Boolean Expressions &Algo.+ Ari.  &Evaluate the truth value of a random Boolean expression consisting of Boolean constants (\texttt{True}, \texttt{False}) and basic Boolean operators (\texttt{and}, \texttt{or} and \texttt{not}). \\

Causal Judgement  &Knowledge  &Given a short story (involving moral, intentional, or counterfactual analysis), determine how a typical person would answer a causal question about the story.\\

Date Understanding  &Knowledge  &Given a small set of sentences about a particular date, answer the provided question (e.g., “The concert was scheduled to be on 06/01/1943, but was delayed by one day to today. What is the date yesterday in \texttt{MM/DD/YYYY}?”).\\

Disambiguation QA &NLU  &Given a sentence with an “ambigious” pronoun, either determine whether the sentence is inherently ambiguous (i.e., the thing that the pronoun refers to cannot be inferred by given information) or, if the pronoun can be implicitly deduced, state the antecedent of the pronoun (i.e., the noun to which the pronoun refers). \\

Dyck Languages  &Algo.+ Ari.  &Predict the sequence of the closing parentheses of a Dyck-4 word without its last few closing parentheses. \\

Formal Fallacies  &Algo.+ Ari.  &Given a context involving a set of statements (generated by one of the argument schemes), determine whether an argument—presented informally—can be logically deduced from the provided context \\

Geometric Shapes &Algo.+ Ari.  &Given a full SVG path element containing multiple commands, determine the geometric shape that would be generated if one were to execute the full path element. \\

Hyperbaton  &NLU  &Given two English-language sentences, determine the one with the correct adjective order. \\

Logical Deduction  &Algo.+ Ari.  &Deduce the order of a sequence of objects based on the clues and information about their spacial relationships and placements. \\

Movie Recommendation  &Knowledge  &Given a list of movies a user might have watched and liked, recommend a new, relevant movie to the user out of the four potential choices user might have. \\

Multi-Step Arithmetic &Algo.+ Ari.  &Solve multi-step equations involving basic arithmetic operations (addition, subtraction, multiplication, and division). \\

Navigate  &Algo.+ Ari.  &Given a series of navigation steps to an agent, determine whether the agent would end up back at its initial starting point. \\

Object Counting  &Algo.+ Ari.  &Given a collection of possessions that a person has along with their quantities (e.g., three pianos, two strawberries, one table, and two watermelons), determine the number of a certain object/item class (e.g., fruits). \\

Penguins in a Table  &Knowledge  &Given a unique table of penguins (and sometimes some new information), answer a question about the attributes of the penguins. \\

Reasoning about Colored Objects  &Algo.+ Ari.  &Given a context, answer a simple question about the color of an object on a surface. \\

Ruin Names &Knowledge  &Given an artist, band, or movie name, identify a one-character edit to the name that changes the meaning of the input and makes it humorous. \\

Salient Translation Error Detection  &NLU  &Given a source sentence written in German and its translation in English, determine the type of translation error that the translated sentence contains. \\

Snarks  &NLU  &Given two nearly-identical sentences, determine which one is sarcastic. \\

Sports Understanding &Knowledge  &Determine whether a factitious sentence related to sports is plausible. \\

Temporal Sequences  &Algo.+ Ari.  &Given a series of events and activities a person has completed in the course of a day, determine what time, during the day, they might have been free to perform another activity. \\

Tracking Shuffled Objects  &Algo.+ Ari.  &Given the initial positions of a set of objects and a series of transformations (namely, pairwise swaps) applied to them, determine the final positions of the objects.\\

Web of Lies  &Algo.+ Ari.  &Evaluate the truth value of a random Boolean function expressed as a natural-language word problem. \\

Word Sorting &Algo.+ Ari.  &Given a list of words, sort them lexicographically. \\
\bottomrule
\end{tabular}
\label{tab:bbh-data}
\end{table*}

\section{LLM-based classifier in \emph{Type Matching}}\label{app:type}
We detail the implementations and provide extended analysis on the alternative in \emph{Type Matching} module: the LLM-based classifier. The proposed classifier employs few-shot examples in the prompt to group the questions based on its \emph{category} and \emph{form}. To implement the LLM-based classifier, we need to ensure the appropriate way of defining the \emph{type} of questions.

\subsection{Defining the \emph{Type} of Questions.}

\begin{figure}
    \centering
    \includegraphics[width=0.48\textwidth]{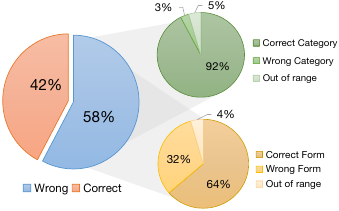}
    \caption{Ratio of wrong cases in task classification.}
    \label{fig:pie1}
\end{figure}

As stated in Section \ref{sec:challenge}, we have collected questions from ten reasoning tasks to set up the mixed-task  scenarios. Those questions cover three categories including arithmetic, commonsense, and symbolic reasoning, and three forms encompassing short-answer, multiple-choice, and yes-or-no questions. Initially, we make a simple attempt to test how well LLMs can identify various tasks (i.e., regarding the question type as task name). We randomly sample one question from each of the ten tasks. For each question, we retain the task name from which it originates so that we obtain ten question-task pairs, which we employ as ICL demonstrations for task classification.
As can be seen from Figure \ref{fig:pie1}, the classification accuracy is only 42\%, which indicate that LLMs are not qualified for distinguishing task  names. Meanwhile, we discover that up to 92\% and 64\% of wrong examples belong to the same category and form as the correct task respectively. We speculate that the underlying reason can be two-fold: on one hand, task names themselves are too abstract for LLMs to well perceive their differences through in-context learning alone. On the other hand, there exist potential similarities and correlations among tasks themselves \citep{zhang-etal-2022-task}. Based on this, we try three schemes for defining the type of questions based on: (i) category; (ii) form; (iii) category and form. 

\subsection{Determining the \emph{Type} of Questions.}

\begin{figure}
    \centering
    \includegraphics[width=0.42\textwidth]{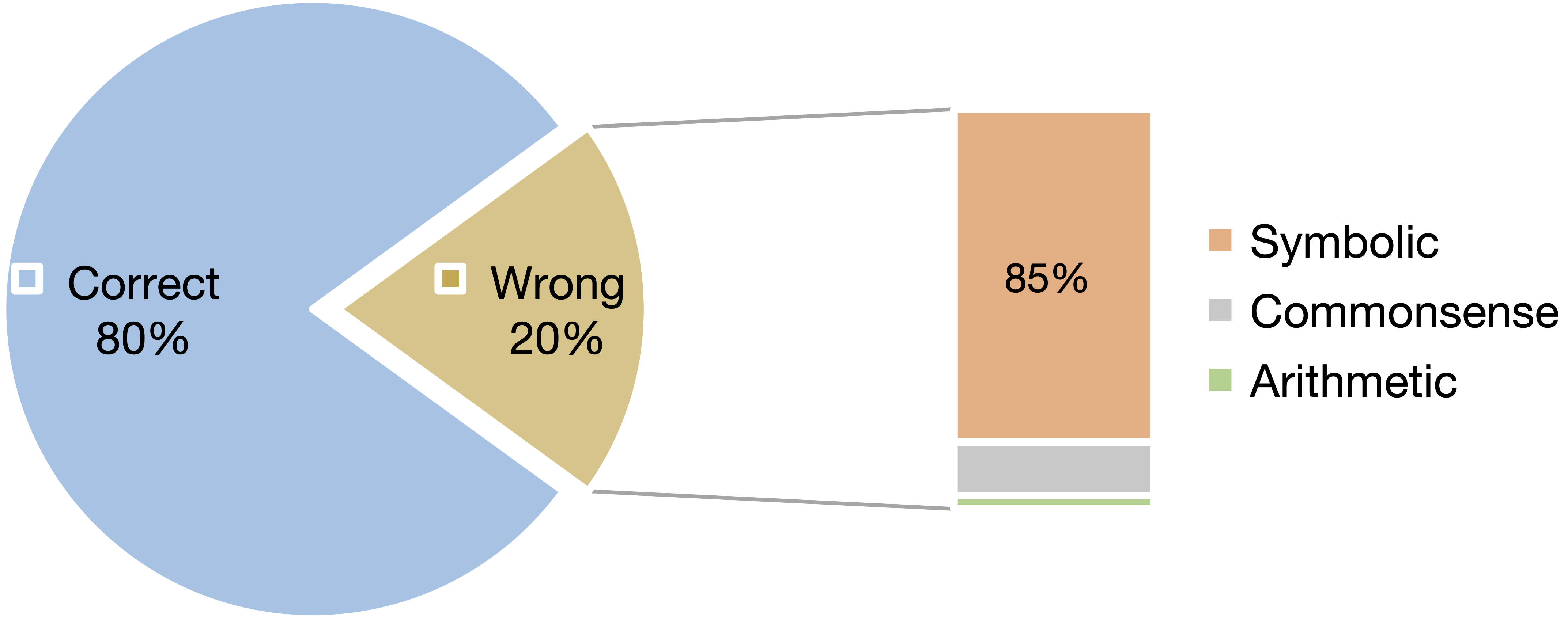}
    \caption{Ratio of wrong cases in category classification, 85\% of wrong cases are from symbolic category.}
    \label{fig:pie2}
\end{figure}

\begin{figure}
    \centering
    \includegraphics[width=0.42\textwidth]{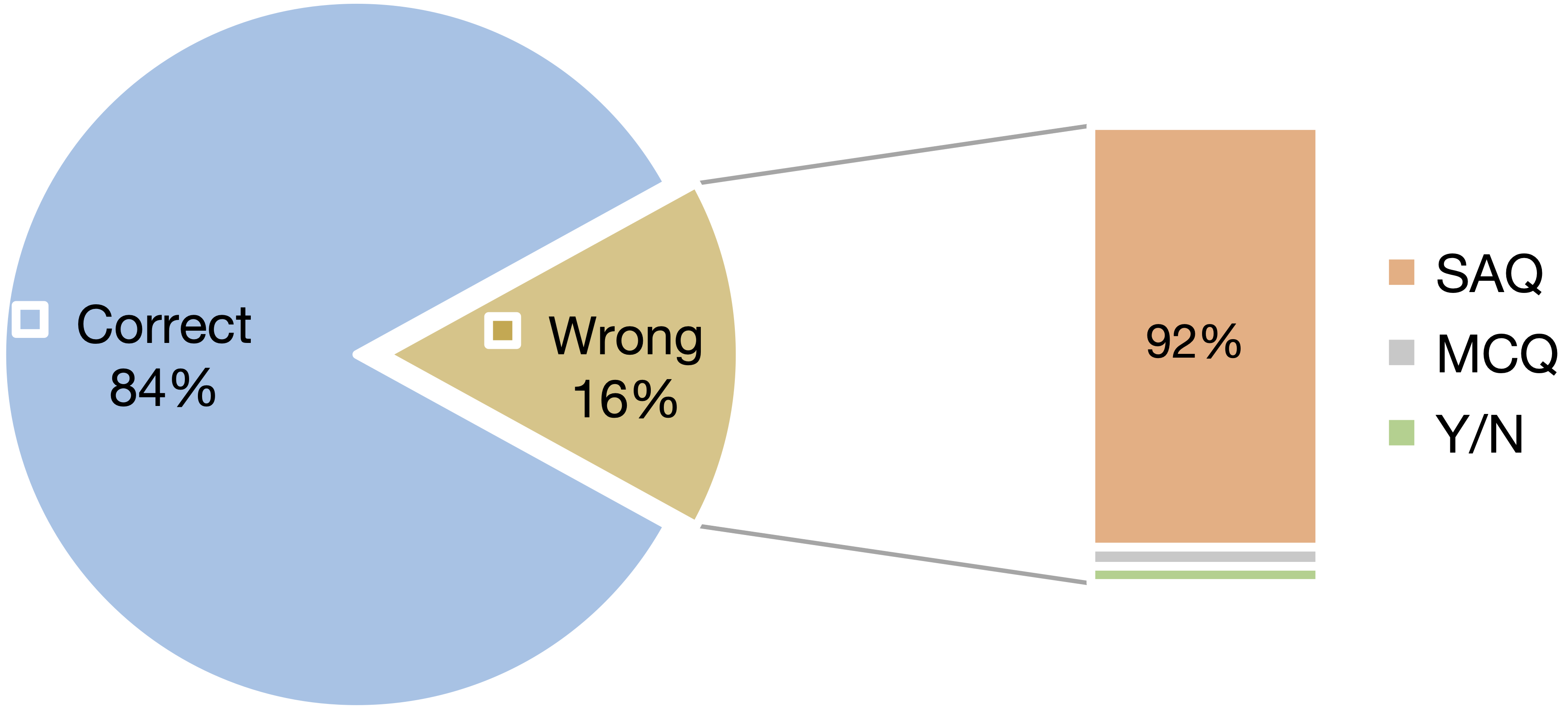}
    \caption{Ratio of wrong cases in form classification, 92\% of wrong cases are from SAQ form.}
    \label{fig:pie3}
\end{figure}

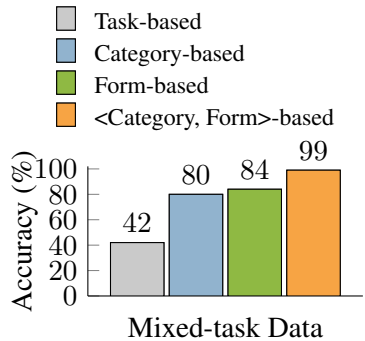
\begin{figure}
\pgfplotsset{width=5.2cm, height=3.3cm}
    \centering
    \begin{tikzpicture}  
        \begin{axis}  
        [  
            ybar,
            ymin=0, ymax=102,
            ytick={0, 20, 40, 60, 80, 100},
            major x tick style = transparent,
            bar width=20pt,
            enlarge x limits=0.2,
            ylabel={Accuracy (\%)},
            symbolic x coords={Mixed-task Data},  
            xtick=data,  
            nodes near coords,  
            legend style={draw=none},
            nodes near coords align={vertical},  
            y label style={at={(axis description cs:-0.145,0.5)},anchor=south},
                axis x line*=bottom,
                axis y line*=left,
        legend cell align=left,
                legend style={
                        at={(0.4,1.2)},
                        anchor=south,
                        column sep=1ex,
                        font=\small,
                }
            ]  
        \addplot[ybar, fill=mygrey,  postaction={}] coordinates {
            (Mixed-task Data, 42)
        };
        \addplot[ybar, fill=myblue,  postaction={}] coordinates {
            (Mixed-task Data, 80)
        };  
        \addplot[ybar, fill=mygreen,  postaction={}] coordinates {
            (Mixed-task Data, 84)
        };  
        \addplot[ybar, fill=myorange,  postaction={}] coordinates {
            (Mixed-task Data, 99)
        };  
        \legend{Task-based, Category-based, Form-based, {\textless Category, Form\textgreater}-based }
        \end{axis}  
    \end{tikzpicture}  
    \caption{Classification accuracy (\%) with different partitioning schemes.}
    \label{fig:partition}
\end{figure}

Since the majority of cases that misidentify task names fall into the same category or form, we compare the classification accuracy with the following three variants of partitioning schemes: (i) Category-based scheme which separates mixed questions into diverse categories; (ii) Form-based scheme which segments data into different answer forms; (iii ) <Category, Form>-based scheme which concurrently takes the two aspects into account. As is shown in Figure \ref{fig:pie2} and \ref{fig:pie3}, we particular group tends to dominate the wrong cases. For instance, 85\% of wrong cases in category classification belong to the symbolic group. We discover that this is because the sampled symbolic group demonstrations do not cover symbolic yes-or-no question, thus hindering LLMs from accurately identifying this missing type. As such, partitioning mixed questions based on both its category and form is a sensible strategy. The results in Figure
 \ref{fig:partition} show that this strategy reaches high accuracy.

Through further experiments, we conclude that defining the type of questions based on its \textbf{category and form} is a sensible strategy, which adequately considers the two major natures of question data and achieves high classification accuracy as well.

\subsection{Constructed Demonstrations for the LLM-based classifier}\label{app:demo}
Table \ref{tab:demo_type} shows the constructed demonstrations for the LLM-based classifier.

\begingroup
\begin{table*}
    \centering
    \caption{
    Constructed demonstrations for type classification. \label{tab:demo_type}
    }
    \vspace{2.8mm}
    \begin{tabular}{p{0.96\linewidth}}
        \toprule
        \textbf{Q:} Bobby had 32 pieces of candy. He ate some pieces of candy. If he has 20 pieces of candy left How many pieces of candy did Bobby eat? \\
\vspace{-1mm}
\textbf{Type:} <arithmetic, short-answer>\\
\vspace{0mm}
\textbf{Q:} The man took paperwork to other people to consult over it, where was he heading? Answer Choices: (A) desk (B) meeting (C) office (D) table (E) work \\
\vspace{-1mm}
\textbf{Type:} <commonsense, multiple-choice>\\
\vspace{0mm}
\textbf{Q:} A coin is heads up. Kristie does not flip the coin. Johnnie flips the coin. Marisa flips the coin. Derick does not flip the coin. Is the coin still heads up? Note that "flip" here means "reverse". \\
\vspace{-1mm}
\textbf{Type:} <symbolic, yes-no> \\
\vspace{0mm}
\textbf{Q:} Take the last letters of each words in "Cruz Wilber Marilu Malik" and concatenate them.\\
\vspace{-1mm}
\textbf{Type:} <symbolic, short-answer>\\
\vspace{0mm}
\textbf{Q:} A company produces 420 units of a particular computer component every month, at a production cost to the company of \$110 per component, and sells all of the components by the end of each month. What is the minimum selling price per component that will guarantee that the yearly profit (revenue from sales minus production costs) will be at least \$626,400 ? Answer Choices: (A) 226 (B) 230 (C) 240 (D) 260 (E) 280\\
\vspace{-1mm}
\textbf{Type:} <arithmetic, multiple-choice>\\
\vspace{0mm}
\textbf{Q:} Was Aristotle a member of the House of Lords?\\
\vspace{-1mm}
\textbf{Type:} <commonsense, yes-no>\\
\bottomrule
    \end{tabular}
\end{table*}
\endgroup

\section{Comparisons of GeM-CoT and existing CoT methods}
Table \ref{tab:cot_methods} demonstrate the comparisons of our proposed GeM-CoT and existing CoT methods in an intuitive and multi-facet way.

\begin{table*}[t]
\centering
\vspace{-4.5mm}
\setlength{\tabcolsep}{10pt}
\small
\caption{Typical CoT techniques (ICL: in-context learning; FT: fine-tuning; KD: knowledge distillation). Segment 1: fine-tuning techniques; Segment 2: in-context learning techniques. To the best of our knowledge, our work is the first to apply CoT prompting to mixed-task scenarios with enjoyable generality and superior performance without additional manual labor. In our work, we focus on in-context learning techniques, eliminating the burden of fine-tuning LLMs. \label{tab:cot_methods}
}
\begin{tabular}{lccccc} 
\toprule
\multirow{2}{*}{\textbf{Model}} & \multirow{2}{*}{\textbf{Training}} & \textbf{Mixed-task} & \textbf{w/o Manual}  & \textbf{w/ Input-related}\\ 
& & \textbf{Scenarios} & \textbf{Labor}  &  \textbf{Info.} \\
\midrule
Fine-tune-CoT \citep{ho2022large} & KD & \ngmark & \okmark  &  \ngmark\\
LoRAHub \citep{huang2023lorahub} & FT & \okmark & \okmark  & \ngmark \\
\midrule
Zero-Shot-CoT~\citep{zero-shot} & ICL & \okmark & \okmark & \ngmark  \\
Few-Shot-CoT~\citep{wei2023chainofthought} & ICL & \ngmark & \ngmark & \okmark\\
Self-Consistency-CoT~\citep{wang2023selfconsistency} & ICL & \ngmark & \ngmark & \okmark  \\
Least-to-Most Prompting~\citep{zhou2023leasttomost} & ICL & \ngmark & \ngmark & \okmark \\
Auto-CoT~\citep{zhang2023automatic} & ICL & \ngmark & \okmark & \okmark  \\
Active Prompt~\citep{diao2023active} & ICL & \ngmark & \ngmark & \okmark  \\
OPRO~\citep{yang2023large} & ICL & \ngmark & \okmark & \ngmark  \\
GeM-CoT (our work) & ICL & \okmark &\okmark & \okmark \\ 
\bottomrule
\end{tabular}
\vspace{-3mm}
\end{table*}

\section{Interpretability: Case Study and Error Analysis}
\subsection{Wrong Type and Correct Answer}
\begin{figure*}[htb]
\centering
\includegraphics[width=1.0\textwidth]{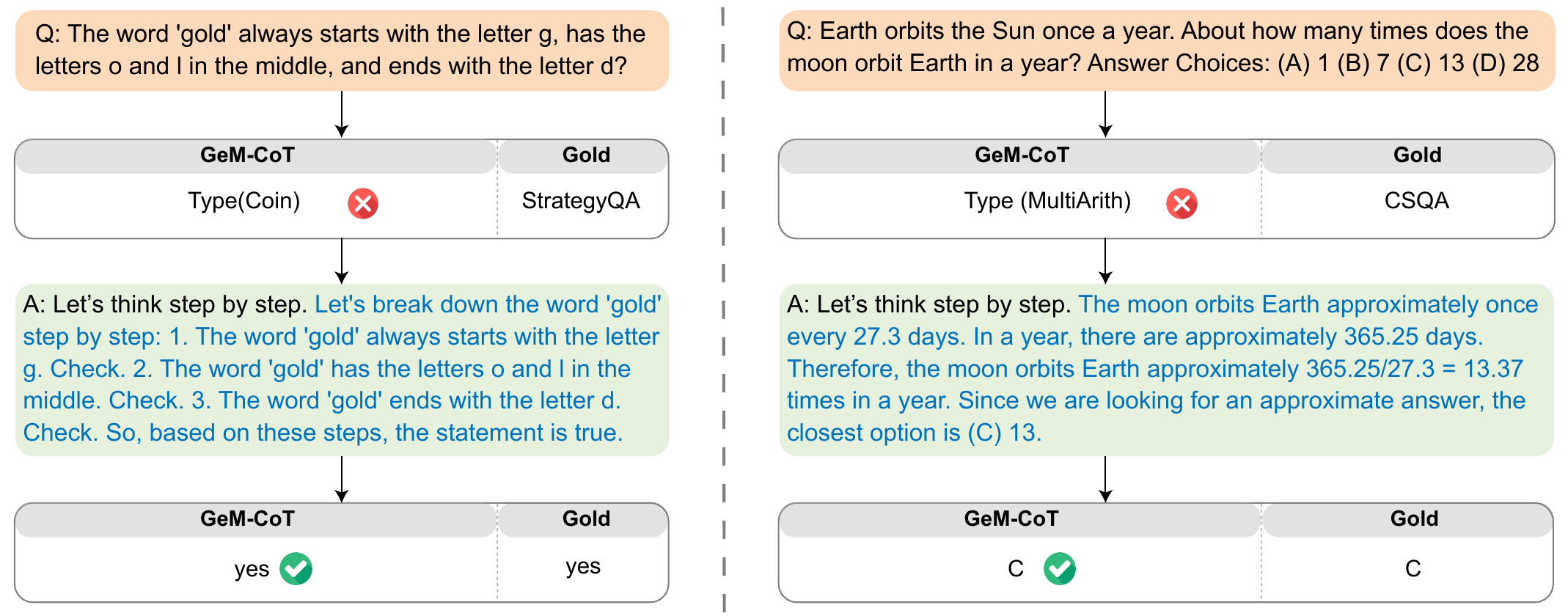}
\caption{Examples from StrategyQA (left) and CSQA (right), in which the type that GeM-CoT identifies is different from the gold type but the final answer from GeM-CoT is correct.}
\label{fig:case}
\end{figure*}

Figure \ref{fig:case} illustrates two examples from StrategyQA and CSQA, in which the type that GeM-CoT identifies differs from the gold type but the final answer from our proposed method is correct. We observe that the proposed \emph{type matching} phase manages to capture the type where the unseen input question is applicable in a more accurate and reasonable way. For instance, the question from StrategyQA (left in Figure \ref{fig:case}) asks whether \emph{the word 'gold' always starts with the letter g, has the letters o and l in the middle, and ends with the letter d}. Although this question belongs to a commonsense question, to answer it would require a process of splitting the word, which has more in common with a symbolic question. Similarly, answering the question from CSQA (right in Figure \ref{fig:case}) necessitates a calculation process, and thus the identified \emph{arithmetic} type leads to more specific and targeted arithmetic reasoning.

\subsection{Wrong Type and Wrong Answer}
\begin{figure*}[htb]
\centering
\includegraphics[width=1.0\textwidth]{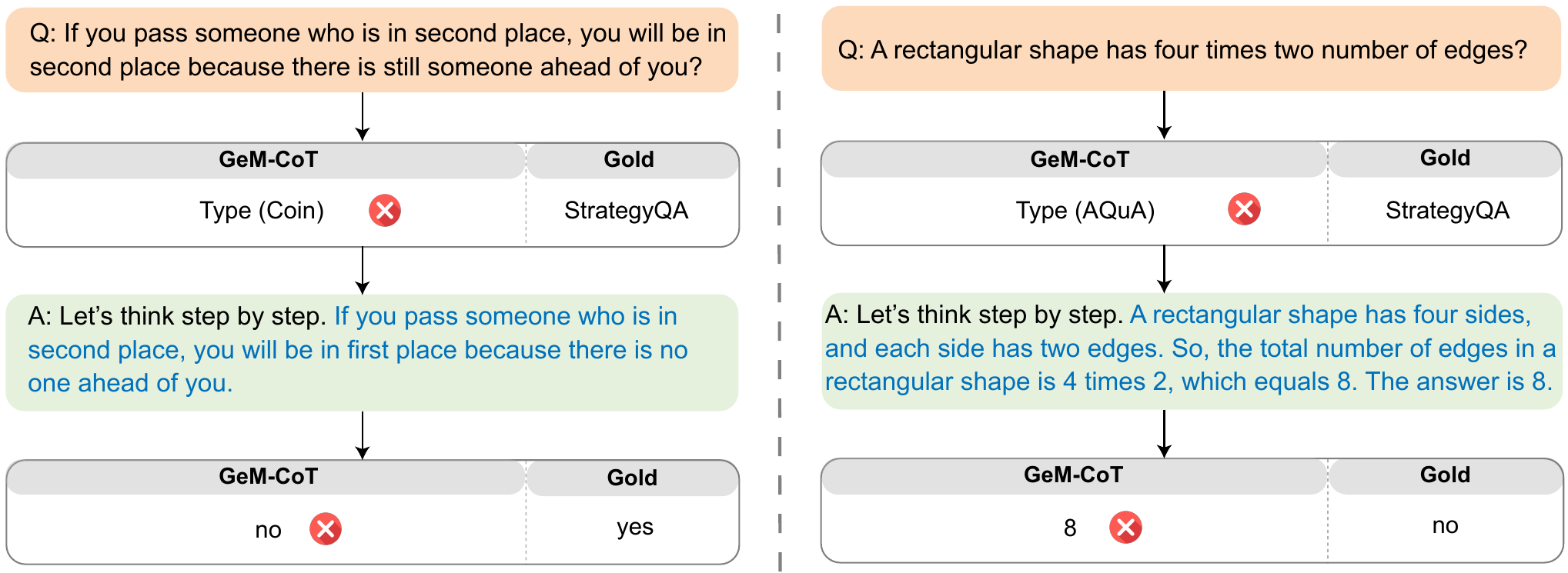}
\caption{Examples from StrategyQA, in which wrongly identified type leads to wrong answer.}
\label{fig:case2}
\end{figure*}

We select two examples from StrategyQA, where GeM-CoT fails but the strategy that provides the model with the gold type succeeds. As is shown in Figure \ref{fig:case2}, we find that some wrongly identified types may result in disastrous reasoning. We analyze that this may be because incorrect ICL demonstrations will disrupt the direction of model inference.

\end{document}